# Instruction-Tuning Large Language Models with Medical Classification Data Improves ICD Coding of German Tumor Diagnoses

*Stefan Lenz[1,4], Lakisha Ortiz Rosario[1], Georg Vollmar[2], Arsenij Ustjanzew[1], Fatma Alickovic[1], Thomas Kindler[2,3], Torsten Panholzer[1]*

[1] Institute of Medical Biostatistics, Epidemiology and Informatics, University Medical Center of the Johannes Gutenberg University Mainz, Mainz, Germany
[2] University Cancer Center of the Johannes Gutenberg University Mainz, Mainz, Germany
[3] Third Department of Medicine, University Medical Center of the Johannes Gutenberg University Mainz, Mainz, Germany
[4] Corresponding author, e-mail: stefan.lenz@uni-mainz.de

ORCIDs: Stefan Lenz https://orcid.org/0000-0001-9135-1743, Lakisha Ortiz Rosario https://orcid.org/0009-0008-9190-4350, Georg Vollmar https://orcid.org/0009-0007-9544-2838, Arsenij Ustjanzew https://orcid.org/0000-0002-1014-4521, Fatma Alickovic https://orcid.org/0009-0001-6885-0269, Thomas Kindler https://orcid.org/0000-0002-0383-0507, Torsten Panholzer https://orcid.org/0000-0002-2808-6120

## Abstract

Accurate coding of tumor diagnoses, with the German Modification of the International Classification of Diseases, 10th Revision (ICD-10-GM), and the International Classification of Diseases for Oncology, Third Edition (ICD-O-3), is essential for structured cancer documentation. Large language models (LLMs) offer opportunities for automating such tasks, but locally deployable, open-weight models often lack sufficient domain-specific coding expertise. This study investigates whether instruction-based fine-tuning on public classification resources improves the coding accuracy of open-weight LLMs for German tumor diagnosis texts. Eight models from the Qwen, Llama, and Mistral families (7–70 billion parameters) were fine-tuned on over 500,000 question-answer pairs based on the ICD-10-GM, ICD-O-3, and German procedure classification. The evaluation uses coded diagnoses from the local tumor documentation system as test data. ICD-10-GM accuracy rose from 1.4-24% to 41-58%, and partial (three-character codes only) accuracy from 31-74% to 73-83%. The accuracy of ICD-O-3 topography coding also improved but started and remained considerably lower. Malformed code outputs dropped to nearly 0% for almost all models. Tumor-diagnosis recognition reached 99%. A data quality analysis showed that incomplete diagnosis descriptions limited achievable test performance. Accuracy correlated positively with model size, but gaps between small and large models narrowed after fine-tuning. The reasoning mode in Qwen3 generally yielded lower performance than fine-tuning and was over 100 times slower. Our findings highlight the potential of leveraging public catalogues to build instruction datasets that improve LLMs in medical documentation tasks. To support reproducibility and further research, we publicly release the complete training dataset and the best-performing fine-tuned model checkpoints.

This version of the article has been accepted for publication, after peer review but is not the Version of Record and does not reflect post acceptance improvements, or any corrections. The Version of Record is available online at: https://doi.org/10.1038/s41598-026-73354-5
Originally submitted under the title: “Unlocking Public Catalogues: Instruction-Tuning LLMs for ICD Coding of German Tumor Diagnoses”

**Keywords**

- Natural Language Processing
- Large Language Models
- Clinical coding
- ICD-10
- ICD-O-3
- German language

**Acknowledgements**

We thank Dr. Meike Ressing for the discussion of selected difficult coding cases in the dataset, and we thank Marco Jeray for providing legal guidance on matters related to data extraction and analysis.

This research has been supported by the Federal Ministry of Research, Technology and Space (BMFTR) in Germany in the project "Digitaler FortschrittsHub Gesundheit – DECIDE" (FKZ 01ZZ2106A) of the German national medical informatics initiative. The project aims to facilitate data transfer and interoperability among different stakeholders in the healthcare sector and to provide decision support based on structured patient information.

## 1. Background

The semantic coding of medical content from unstructured text sources is an important part of structured information extraction in medicine. Currently, correctly coding tumor diagnoses requires skilled personnel familiar with the specific coding requirements of tumor documentation. As this is a purely text-based task, automating this process has been in the focus of natural language processing (NLP) research [1]. Automated support for this task could substantially reduce the manual workload associated with tumor documentation. Large language models (LLMs) are a promising NLP approach to this complex problem [2]. However, further improvements are needed to make these tools viable for practical use in German-language tumor documentation [3].

Whenever possible, structured information extraction approaches should be based on established classification systems to enhance data reusability. One of the most widely used medical classification systems is the International Classification of Diseases (ICD) [4]. For the coding of tumor diagnoses, there is the International Classification of Diseases for Oncology (ICD-O). The use of both ICD-10 ($10^{th}$ revision of the ICD catalogue) and ICD-O for documenting tumor diagnoses and reporting them to the cancer registries is mandated by law for the tumor documentation in Germany [5].

The ICD-10 codes for tumor diagnoses are single codes in the range of C00.0-D48.9 [4]. The ICD-O code consists of two parts [6]. The first part is the ICD-O topography code, which encodes the localization in the body and assumes values in the range of C00.0-C80.9. Secondly, there is the 5-digit ICD-O morphology code, which requires an examination of tumor tissue to be determined and provides greater detail about the histological properties of the tumor. While there is a substantial semantic overlap and great similarity between ICD-O and ICD-10, there are many cases where the information about a tumor diagnosis cannot be translated from one coding system to the other without losing information. As a result, both systems are used alongside each other rather than one replacing the other. This study evaluates only the topography component of ICD-O, as ICD-O topography codes and ICD-10 codes are both used to code the primary tumor in the German cancer registry dataset [7]. In contrast, ICD-O morphology codes are assigned based on histopathological assessment and are recorded in separate entries that correspond to pathological examinations.

Assigning ICD codes to textual diagnoses is a difficult task and it has been shown that even powerful LLMs can struggle with it [8]. Adapting pretrained models to the specific structure and terminology of medical coding systems may be necessary to achieve reliable coding performance. However, task-specific adaptation depends on the availability of suitable training data. Due to privacy restrictions, clinical text data for training are scarce [9]. This is especially challenging for languages other than English, such as German, for which publicly accessible clinical text datasets with code annotations are limited [10]. Improving the utilization of publicly available medical classification catalogues may therefore provide an alternative to relying on sensitive clinical data. Although the terminology used in classification systems can differ from clinical language [11], catalogue-based training data have the advantage that they are not tied to the jargon and documentation practices of a single institution and therefore lack site-specific linguistic dependencies.

Modern LLMs are pretrained on very large text corpora, but the extent to which specific medical coding systems are represented in these data is generally uncertain. It therefore remains an open question whether coding performance can be improved by providing models with targeted exposure to relevant coding knowledge beyond what they may have encountered during pretraining. Both additional medical domain adaptation and task-specific fine-tuning have been shown to improve ICD coding performance [9,12].

Instruction-based fine-tuning has improved performance for automated clinical information extraction and coding tasks, including named entity recognition, relation extraction, and ICD coding from clinical text, although room for further improvement remains [12–15]. In contrast to approaches that derive instruction-tuning examples from annotated clinical data [12–14] or biomedical NLP datasets [15], our approach constructs task-specific training examples directly from publicly available medical coding catalogues. Here, we represent the coding task as an instruction based on the textual catalogue entry with the corresponding code as a predefined response. This allows us to create training examples that convey both coding-system knowledge and the expected response format without relying on annotated clinical texts. Using exclusively public training data facilitates the release and reuse of the resulting fine-tuned models by avoiding privacy concerns and the effort associated with anonymizing sensitive clinical data.

Therefore, this study investigates whether publicly available ICD-10-GM and ICD-O catalogues can be transformed directly into instruction-tuning data to improve the coding performance of LLMs for German tumor diagnoses. To assess whether the acquired coding knowledge generalizes to real-world clinical descriptions, we evaluate the fine-tuned models using actual diagnosis texts from the local tumor documentation system at the University Medical Center Mainz.

## 2. Methods

### *2.1. A training dataset of instructions constructed from public catalogues*

Instruction-tuning of large language models is generally performed using question-answer pairs. For improving the documentation capabilities of the models, we designed question-answer pairs that encapsulate knowledge about coding tumor diagnoses. These training examples were derived deterministically from fixed German-language instruction templates that were formulated separately for each task. The templates were created manually without the use of LLMs. The different formulations within a task were intended to be semantically equivalent but to expose the models to different phrasings of the same request. Each template contained a placeholder for a textual catalogue entry. For each task, every eligible catalogue entry was combined with all instruction formulations defined for that task by inserting its textual description into the respective placeholder. The corresponding answer was obtained directly from the code or class label provided by the catalogue. The question-answer pairs were created using the following public catalogues used for medical documentation in Germany:

- Alpha-ID (alphabetical index of the German version of the International Classification of Diseases (ICD-10-GM), 2024 version) [16]
- ICD-O-3 (International Classification of Diseases for Oncology, German version, second revision, 2019) [6]
- OPS (German classification of medical procedures, 2024 version) [17]

The variant of the Alpha-ID catalogue used contains 13,600 tumor diagnosis descriptions in the chapter concerning neoplasms (code range C00-D48), corresponding to a total of 809 distinct codes. These codes and descriptions were all used to construct the training dataset. The same number of non-tumor diagnosis descriptions were sampled from the Alpha-ID catalogue and included to have an even distribution for training the tumor diagnosis recognition task. From the ICD-O-3 catalogue, we extracted and included all available data for the training. This led to 1,338 localization descriptions with 400 distinct ICD-O topography codes and 2,804 tumor diagnosis descriptions with 1,145 distinct ICD-O morphology codes used for training.

It has been shown that increasing the diversity of tasks in instruction-tuning datasets can lead to better training results [18]. Accordingly, the training dataset comprised several related tasks, primarily ICD-10 and ICD-O coding as well as the distinction of tumor diagnoses from non-tumor diagnoses and medical procedures. Within each task, multiple semantically equivalent instruction formulations were used to expose the models to different phrasings of the same request. The textual descriptions inserted into the instruction templates were taken directly from the respective catalogues and therefore retained the terminology and linguistic variation present in these source data. For example, three Alpha-ID entries associated with the same ICD-10-GM code C34.2 are “Lungenkarzinom des Mittellappens” (“lung carcinoma of the middle lobe”), “Neuroendokriner Tumor des Mittellappenbronchus” (“neuroendocrine tumor of the middle lobe bronchus”), and “Pulmonales Blastom des Lobus medius pulmonis” (“pulmonary blastoma of the middle lobe of the lung”, with the anatomical site expressed using Latin terminology).

The tumor diagnosis recognition tasks additionally included non-tumor diagnoses and medical procedures as negative examples. In addition to a sample of non-tumor diagnoses from the Alpha-ID catalogue, all the main categories of procedures from the OPS catalogue were included as additional negative examples to sharpen the understanding of the concept of tumor diagnoses. In practice, this distinction is important because operations and procedures may be related to tumor diagnoses and some of them imply a tumor diagnosis, but they do not constitute diagnoses themselves. Procedures should therefore not be reported as diagnoses in the documentation.

**Table 1**. Distribution of question-answer pairs in the training dataset.

| **Questions asking for …** | **No. of formulations** | **No. of questions** | **Proportion of questions** |
|---|---|---|---|
| ICD-10 codes (only tumor diagnoses) | 12 | 163,200 | 31.5% |
| Recognizing tumor diagnoses from the Alpha-ID catalogue, yes/no answers (50% tumor and non-tumor diagnoses) | 6 | 163,200 | 31.5% |
| Recognizing tumor diagnoses from the Alpha-ID catalogue, answers with yes/no and ICD-10 code (50% tumor and non-tumor diagnoses) | 4 | 108,800 | 21.0% |
| OPS main categories | 10 | 29,640 | 5.7% |
| Recognizing operations/procedures from the OPS catalogue as distinct from tumor diagnosis | 4 | 11,856 | 2.3% |
| ICD-O morphology codes from the ICD-O-3 catalogue | 10 | 28,040 | 5.4% |
| ICD-O topography codes from the ICD-O-3 catalogue | 10 | 13,380 | 2.6% |
| **Total** | | **518,116** | **100%** |

An overview of the types of question-answer pairs that are included in the dataset can be seen in Table 1. The instructions are phrased in German. Translated into English, example questions include “What is the ICD-10 code for [diagnosis]?”, or “Which ICD-O topography code fits [diagnosis]?”, with the corresponding code as the answer. Similarly, questions related to tumor diagnosis recognition are variations of “Is ‘[…]’ a tumor diagnosis?”, to which the given answer is either “Yes” or “No”. The same phrasings of these questions were also used later for evaluating fine-tuned models. To increase the task variety in the training data, we added similarly phrased questions asking for OPS main category codes and ICD-O morphology codes to the training data. Additionally, we included questions combining tumor diagnosis recognition and coding, e. g., by asking “Which ICD-10 code belongs to the diagnosis ‘[…]’, and is it a tumor diagnosis?”.

Each type of question is provided in a number of different formulations with identical meaning (see second column in Table 1). This way, the model is trained to respond to a wider range of prompt formulations in the desired manner.

The constructed answers are kept short to instruct the model to return short answers, which are faster to generate and easier to process than longer answers.

Although morphology coding was not part of the evaluation, morphology examples were included in the instruction-tuning dataset solely to help the models distinguish ICD-O morphology from topography and learn the structure of ICD-O coding.

### 2.2. *Real-world test data from the tumor documentation system*

To evaluate the effect of training the models on the data assembled from the catalogues, we use a real-world test data set from the local tumor documentation system of the University Cancer Center at the University Medical Center Mainz. We only used the short diagnosis descriptions together with the corresponding ICD-10 and ICD-O topography codes from the table describing the primary tumor in the local tumor documentation database. These texts serve as real-world counterparts to the catalogue descriptions that were used for training. Additional structured or textual clinical information from the tumor documentation system was not included. This kept the training and test data comparable and was consistent with the study objective of coding short textual diagnosis descriptions of the kind also found in physicians’ letters to describe a tumor diagnosis.

Although ICD-O morphology coding data were available in the underlying oncology database, they were not included in the test set. We decided to focus solely on ICD-10 and ICD-O topography coding because the morphology information is stored separately in structured histopathology entries. Integrating it would have required additional linkage and reconciliation across potentially multiple examinations associated with the same tumor. More importantly, an initial assessment indicated that the short diagnosis descriptions given for the primary tumor input generally lacked the histopathological detail required to derive fine-grained morphology codes reliably.

The study was conducted in accordance with the principles of the Declaration of Helsinki and all relevant guidelines and regulations. The responsible ethics commission (Ethikkommission der Landesärztekammer Rheinland-Pfalz) stated on 06 September 2024 that formal ethics approval was not required for this study because only short diagnosis descriptions and codes without identifying information were used. The institutional data protection office confirmed that the use of these data for research purposes without informed consent, including the preparation of the anonymized dataset and its analysis within the hospital network, was permissible according to § 37(1)(3) State Hospital Act of Rhineland-Palatinate (Landeskrankenhausgesetz Rheinland-Pfalz).

The test data were extracted on June 27, 2024. We used only recent diagnoses from the years 2023 and 2024. This way, potential differences with older versions of the ICD-10-GM catalogue, which is updated yearly, can be avoided. Restricting the number of diagnoses this way also keeps the computation time for the evaluation manageable. Furthermore, duplicated textual descriptions were filtered out.

The resulting data used for testing the models (test data) comprised 2024 unique textual descriptions of diagnoses in total, which are labeled with the corresponding ICD-10 code and the ICD-O topography code. The distribution of codes in the test dataset was uneven, reflecting differences in the frequency of tumor entities in the underlying clinical data. For ICD-10, the most frequent three-character categories were C44 ("Other malignant neoplasms of skin", 19.42%), C43 ("Malignant melanoma of skin", 14.23%), and C34 ("Malignant neoplasm of bronchus and lung", 5.34%) whereas for ICD-O topography, C44 ("Skin") accounted for 36.51% of cases, and C34 ("Bronchus and lung") for 5.53%. At the same time, the remaining cases were distributed across a broad range of less frequent categories. The median number of cases per category was 8 for ICD-10 and 12 for ICD-O topography. Overall, the test dataset covered 87 distinct three-character ICD-10 categories across 136 categories in the ICD-10 chapter concerning neoplasms (C00–D48). It also covered 60 distinct ICD-O topography categories out of the 70 categories spanning C00–C80 in the ICD-O-3 classification. The complete code distributions are provided in Supplementary Table S1 and Table S2.

To quantify the textual overlap between the catalogue-based training data and the primary test dataset, we compared the diagnosis descriptions in both datasets. An exact textual match with an Alpha-ID description used for fine-tuning was found for 9.73% of the test descriptions. In addition, we performed a less restrictive comparison by converting the descriptions into bag-of-words representations after stemming with the German Snowball stemmer implemented in NLTK [19]. Using this normalized representation, 10.92% of the test descriptions had a corresponding representation in the training data. The bag-of-words comparison disregards word order and, through stemming, reduces inflected word forms to common stems, and therefore captures a broader degree of lexical overlap than exact string matching.

To complement this primary evaluation and assess whether the observed effects also extend to data from another institution with a different case mix, we performed an additional exploratory evaluation using the publicly available HANCOCK dataset, which also contains ICD-10-GM codes together with short diagnosis descriptions in German [20]. Due to its limited size, this analysis is presented separately in Appendix B, while the main analysis focuses on the test dataset from University Medical Center Mainz.

The three different tasks of assigning an ICD-10 code, assigning an ICD-O code, and recognizing the description as a description of a tumor diagnosis were evaluated separately with three different sets of evaluation questions that are presented to the models. The questions for evaluating the models were constructed in the same manner as the training questions.

### 2.3. *Data quality assessment*

The available data were extracted from only three columns of the comprehensive database of the local tumor documentation system, representing a narrow subset of the full patient information. In routine use, documenters enter short free-text descriptions of tumor diagnoses in the context of extensive additional clinical information. As a result, these descriptions are not primarily designed to capture all details required for code derivation. Nevertheless, in most cases the short description is sufficient to derive a code. To account for the issue of incomplete information, we carried out an analysis of the quality of the descriptions with respect to their suitability for coding.

We identified two main reasons why codes cannot be derived from the textual description: First, if the tumor behavior (benign/malignant/uncertain) is not specified, an appropriate ICD-10 code cannot be assigned. Second, a missing anatomical localization can hinder accurate ICD-O topography coding.

An example for missing information about the tumor behavior in the data is the description "Nierentumor der rechten Niere" (English: renal tumor of the right kidney), which is coded as C64 in the data. The localization is described in detail here, but the behavior is unclear from the description alone. Therefore, several ICD-10 codes could match this description: C64 (if malign), D30.0 (if benign), or D41.0 (if uncertain behavior).

An example where the ICD-O localization cannot be derived fully from the text is the diagnosis description "DLBCL Stadium IIIA". DLBCL (diffuse large B-cell lymphoma) can be readily mapped to the ICD-10 code C83.3. However, the localization cannot be derived from this description alone. Lymphomas, although primarily arising in lymph nodes, may also occur in extranodal sites containing lymphatic cells, such as the gastrointestinal tract, the central nervous system, or even bone [21]. Without explicit mention of the primary site, an accurate ICD-O topography code cannot be assigned.

Due to such cases of incomplete information as illustrated in the examples above, perfect model performance (i.e., 100% accuracy) cannot be expected for ICD-10 coding, ICD-O coding, or tumor diagnosis recognition. To estimate the theoretical upper bound of achievable performance, a data quality analysis was conducted. For this purpose, a random sample of 100 cases (4.94% of the dataset) was selected and annotated independently by two reviewers. The annotations addressed whether the code specified for the diagnosis could be derived from the tumor diagnosis text and classified the diagnosis coding using the categories fully/partially/not derivable. A code was considered fully derivable if the exact code could be determined, and partially derivable if at least the three-character category could be inferred. This classification was performed separately for ICD-10 and ICD-O. In addition, the reviewers documented the reasons for unclear mappings (missing tumor behavior/missing localization/other).

To evaluate the consistency of these annotations, inter-rater agreement was calculated. Cohen's kappa (calculated with R package 'irr', version 0.84.1) indicated substantial agreement among the two annotators, with values of 0.63 for ICD-10 code derivability, 0.68 for ICD-O localization derivability, and 0.70 for the attribution of reasons for unclear mappings. After a joint discussion of the two reviewers and a third researcher, cases that remained uncertain were referred to an experienced specialist. Finally, a consensus version was established.

Table 2 shows the consensus results of the quality assessment for the derivability of the codes. In four cases for the ICD-10 and two of the ICD-O codes, there was a complete mismatch between code and text, i.e., the given code did not fit the textual description. In Table 2, these cases were counted as "not derivable" along with instances where multiple codes fit the text. Overall, a considerable proportion of cases were only partially or not derivable, primarily due to missing localization (31 cases) or non-derivable tumor behavior (3 cases).

**Table 2.** Results of data quality assessment on 100 randomly selected diagnosis descriptions (95% confidence intervals in brackets). "Partially derivable": three-character category can be inferred but not the exact code; "Not derivable": multiple possible codes in different three-character categories or mismatch between text and code.

| Code | Fully derivable (%) | Partially derivable (%) | Not derivable (%) | Total (%) |
|---|---|---|---|---|
| **ICD-10** | 70 (60–79) | 19 (12–28) | 11 (6–19) | 100 |
| **ICD-O** | 65 (55–74) | 26 (18–36) | 9 (4–16) | 100 |

In conclusion, the data quality analysis indicates that the attainable accuracy is bounded at approximately 70% for complete ICD-10-GM code prediction and 65% for exact ICD-O localization. Likewise, the upper bounds for the partial accuracy (prediction of the three-character category) lie around 89% for ICD-10-GM and 91% for ICD-O.

### 2.4. *Model selection*

For this evaluation, we selected eight different LLMs from the "Llama" model family from Meta, the "Qwen" model family from Alibaba, and two models from Mistral AI, ranging from 7 billion model parameters (suffix "7B") to 70 billion model parameters (suffix "70B"). An overview of the models can be found in Table 3.

**Table 3.** Overview of evaluated models with model size (number of parameters), quantization, number of GPUs (Nvidia A40), and fine-tuning time. The models are sorted by size, defined as the number of learned parameters (weights and biases), as specified in their respective model cards on HuggingFace.

| **Model name** | **No. of parameters / $10^9$** | **Quantization (fine-tuning / test)** | **No. of GPUs (fine-tuning / test)** | **Fine-tuning time (hours)** | **Model card** |
|---|---|---|---|---|---|
| Qwen2.5-7B | 7.61 | No / No | 1 / 1 | 102 | [22] |
| Llama 3.1 8B | 8.03 | No / No | 1 / 1 | 96 | [23] |
| Qwen3-8B | 8.19 | No / No | 1 / 1 | 110 | [24] |
| Mistral NeMo | 12.2 | No / 8-bit | 2 / 1 | 118 | [25] |
| Qwen3-14B | 14.8 | 8-bit / 8-bit | 1 / 1 | 86 | [26] |
| Qwen3-32B | 32.8 | 8-bit / 8-bit | 2 / 1 | 179 | [27] |
| Mixtral 8x7B | 46.7 | 4-bit / 4-bit | 2 / 2 | 315 | [28] |
| Llama 3.3 70B | 70.6 | 4-bit / 4-bit | 2 / 2 | 508 | [29] |

While all these models have primarily been trained on English texts, they officially support multiple languages, including German. We have already tested some of these models for ICD-10 classification and found the results promising but with room for improvement [3]. To assess the generalizability of the results, we included models from different model families and vendors here. Moreover, we used models of different sizes to examine how the model size affects the baseline performance of the models and the potential to improve the models via fine-tuning.

### 2.5. *Training and evaluation*

In all cases, instruction-tuned models were used, i.e., models that had been fine-tuned on question-answer pairs to enhance their conversational abilities. Building on this, additional instruction-based fine-tuning was attempted with the training dataset of tailored question-answer pairs, aiming at enhancing the ICD coding expertise of the models. The scripts for implementing the fine-tuning and the evaluation were developed using Python (version 3.12) and the 'transformers' package [30].

The fine-tuning was performed using Low-Rank Adaptation (LoRA), a parameter-efficient technique that updates only low-rank matrices added to specific layers of the models. This technique significantly reduces the computational and memory requirements compared to a full fine-tuning of all model weights [31]. The LoRA adapters were applied to the query and value projection layers of the attention modules. This choice was motivated by the original LoRA study, which found that jointly adapting the query and value projections provided a favorable allocation of adaptation capacity under a fixed parameter budget [31]. The same target modules were used for all models to avoid differences in adaptation capacity arising from architecture-specific module selection. The adapter-specific hyperparameters were within the range of configurations evaluated in the original LoRA study [31]: A moderately low rank of r=8 was selected as a compromise between adaptation capacity and memory efficiency. This choice is consistent with evidence that downstream adaptation of pretrained language models often occurs in a low-dimensional subspace and that parameter-efficient adaptation can retain strong performance under restricted parameter budgets [32]. A correspondingly typical scaling factor of $\alpha$=16 (=2r) was chosen [31,33]. A dropout probability of 0.1 was applied as a moderate regularization measure to reduce the risk of overfitting to the wording of the training data from the medical classifications, which differed from the real-world evaluation data. A weight decay of 0.01 was used as an additional regularization measure. The batch size per device was set to 6 due to memory constraints. Given the small batch size, which may lead to higher gradient variance, a more conservative learning rate of $5\times10^{-5}$ was chosen. Due to the extensive computational cost and training time, no systematic optimization of these hyperparameters was conducted.

During training the loss values and the gradients were monitored. For the models Qwen3-14B and Qwen3-32B, NaN values for the gradients occurred in training epoch 3 when training with a reduced batch size of 2 to fit the models on a single GPU. In these cases, training was resumed from the last valid checkpoint in epoch 2. For Qwen3-14B, resuming the training was sufficient to get a stable result, while the instability persisted with Qwen3-32B. Subsequently, two GPUs were used to train Qwen3-32B, but a reduced learning rate of $10^{-5}$ and gradient norm clipping with a maximum norm of 1.0 were applied.

The textual answers returned from the trained models were compared to the labels and converted to truth values with simple regular expression matching. The models were evaluated before and after each of the five fine-tuning epochs. For a direct comparison to the training data, the performance on the training data from the catalogue was evaluated in the same manner.

The primary evaluation of the base models used zero-shot prompting, with each diagnosis description presented independently and without additional example cases in the prompt. This setting was chosen to reflect the intended use case in tumor documentation, where codes should be generated directly and efficiently from individual diagnosis descriptions without requiring additional context examples for each query. To assess whether this choice underestimated the coding capabilities of the base models, we additionally evaluated three-shot and five-shot prompting for ICD-10 and ICD-O topography coding. For each test instance, the demonstration examples were randomly selected from the Alpha-ID catalogue for ICD-10 coding and from the ICD-O-3 catalogue for ICD-O topography coding. The same randomly selected examples were used for all models for a given test instance to ensure comparability. Apart from the inclusion of these demonstration examples, the same evaluation procedure and performance metrics as in the zero-shot setting were used.

While the base models generally tended to give longer answers, the fine-tuned models returned short answers as they were trained to do this with the question-answer pairs. To shorten the answers of the base models, we added a sentence in the prompt that guides the models to also answer shortly. Answer generation was limited to 20 tokens for all models. Responses that did not follow the instruction to answer briefly could therefore be truncated before the relevant code was generated. The additional instruction reduced the amount of output and increased the success of parsing the answer to a well-defined code. The values of the metrics for the base models are based on these modified prompts to alleviate this problem and make a fairer comparison possible.

Accuracy (complete code match) and partial accuracy (three-character codes match) were calculated among valid answers only, i.e., answers that could be mapped via regular expression matching. Invalid outputs were reported separately. This way of measuring the coding performance was chosen because, in a practical coding application, an incorrect but well-formed code may be more problematic than an output that cannot be parsed and can therefore be flagged for review. This approach was also intended to make the comparison between base and fine-tuned models fairer as the answers of base models that did not adhere to the instruction to answer briefly could therefore be truncated before providing a correct code.

A property unique to the Qwen3 models is their ability to operate in "thinking mode" as well as in "non-thinking mode" [34]. In thinking mode, they generate intermediate reasoning steps before providing an answer, as opposed to responding directly. We used this property to examine the extent to which the reasoning capabilities of LLMs can improve performance in the specific task of ICD coding, and to estimate the additional computational cost introduced by the required reasoning. The probabilistic token outputs from the models were not further sampled for the base models (temperature 0) to avoid the introduction of additional noise. For the thinking process, sampling was required and we followed the recommendation from the Hugging Face model cards of the three Qwen3 models ([24,26,27]) for setting the sampling parameters. Accordingly, we used a temperature of 0.6 and a nucleus-sampling probability threshold (TopP) of 0.95. The top-k cutoff (TopK) was set to 20, restricting sampling to the 20 tokens with the highest probabilities. The min-p threshold (MinP), which filters tokens relative to the probability of the most likely token, was set to zero.

## 3. Results

To provide a brief overview of the main findings, fine-tuning substantially improved ICD-10 and ICD-O topography coding and also increased tumor diagnosis recognition. The strongest gains occurred during the first training epoch. The results are presented in four parts: An initial analysis of the training loss indicates convergence of model training. A closer look on the performance before and after fine-tuning details the performance achieved by the base models and the improvements that were achieved through fine-tuning. We then further examine the association between model size and coding performance and could see that larger models generally retained a performance advantage, although fine-tuning reduced the differences between model sizes. Finally, we evaluate the effect of the Qwen3 thinking mode on accuracy, output validity, and response time. Here we could see that the Qwen3 thinking mode did not improve exact coding accuracy but substantially increased response time.

### 3.1. *Training dynamics*

The loss curves in Figure 1 were smooth and dropped rapidly, especially in the first training epoch. In subsequent training epochs, the loss continued to decline on the training data but at a much slower rate. Apart from the instabilities observed for Qwen3-14B and Qwen3-32B, the models behaved quite similarly in the training.

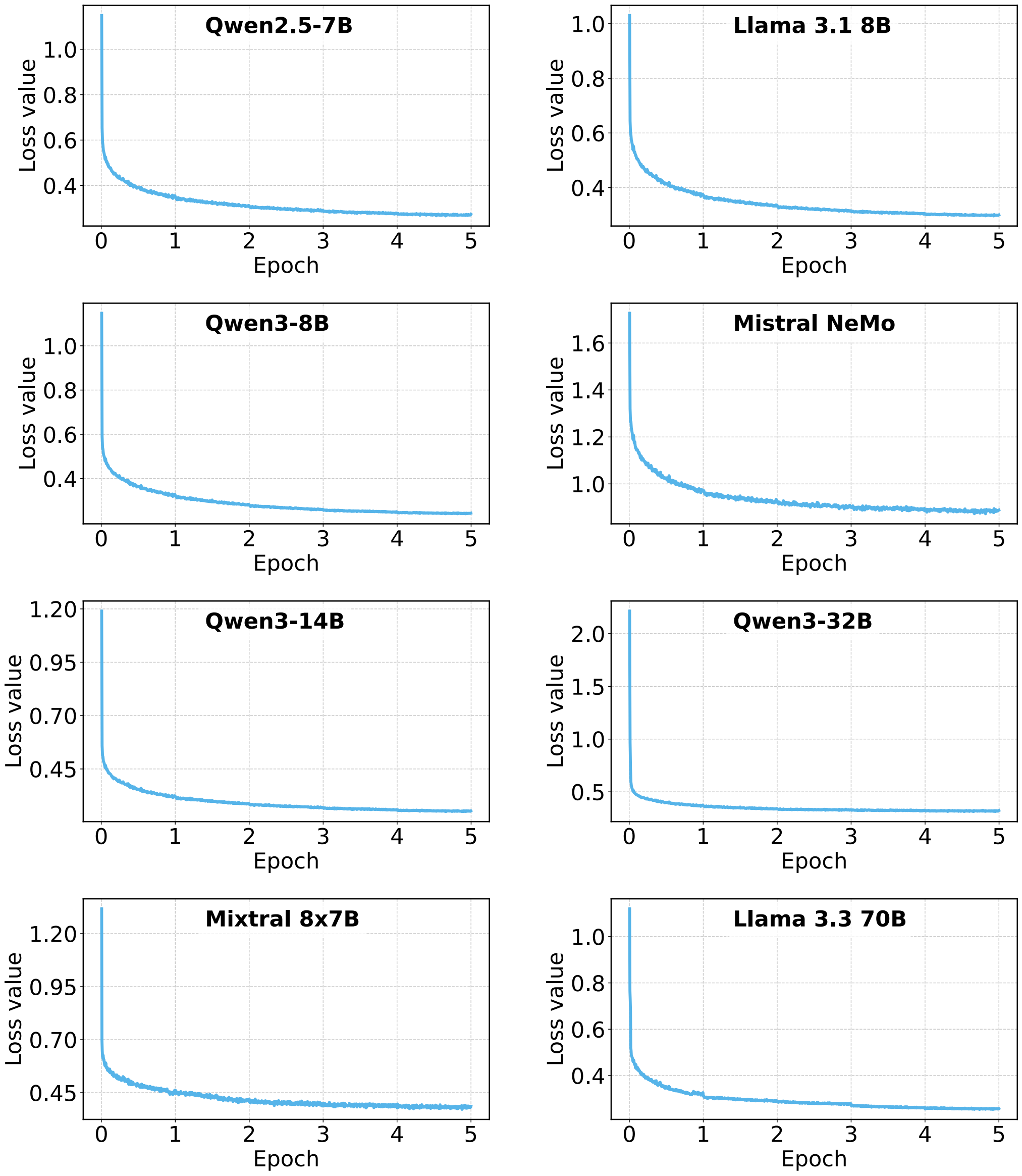

**Figure 1.** Loss curves during instruction-based fine-tuning. The loss is the token-level cross-entropy loss for next-token prediction across the complete question-answer sequences. It is averaged across the full training dataset, which contains question-answer pairs for all tasks.

### 3.2. *Performance improvements through fine-tuning*

Figure 2 shows the performance of the models on the training and test datasets. Almost all models improved substantially after the first training epoch. The performance on the training data also improved further for ICD-10 and ICD-O coding after the first training epoch. However, there were no substantial gains on the test data after the first epoch for ICD-10 classification, and in some cases, there was even a slight deterioration.

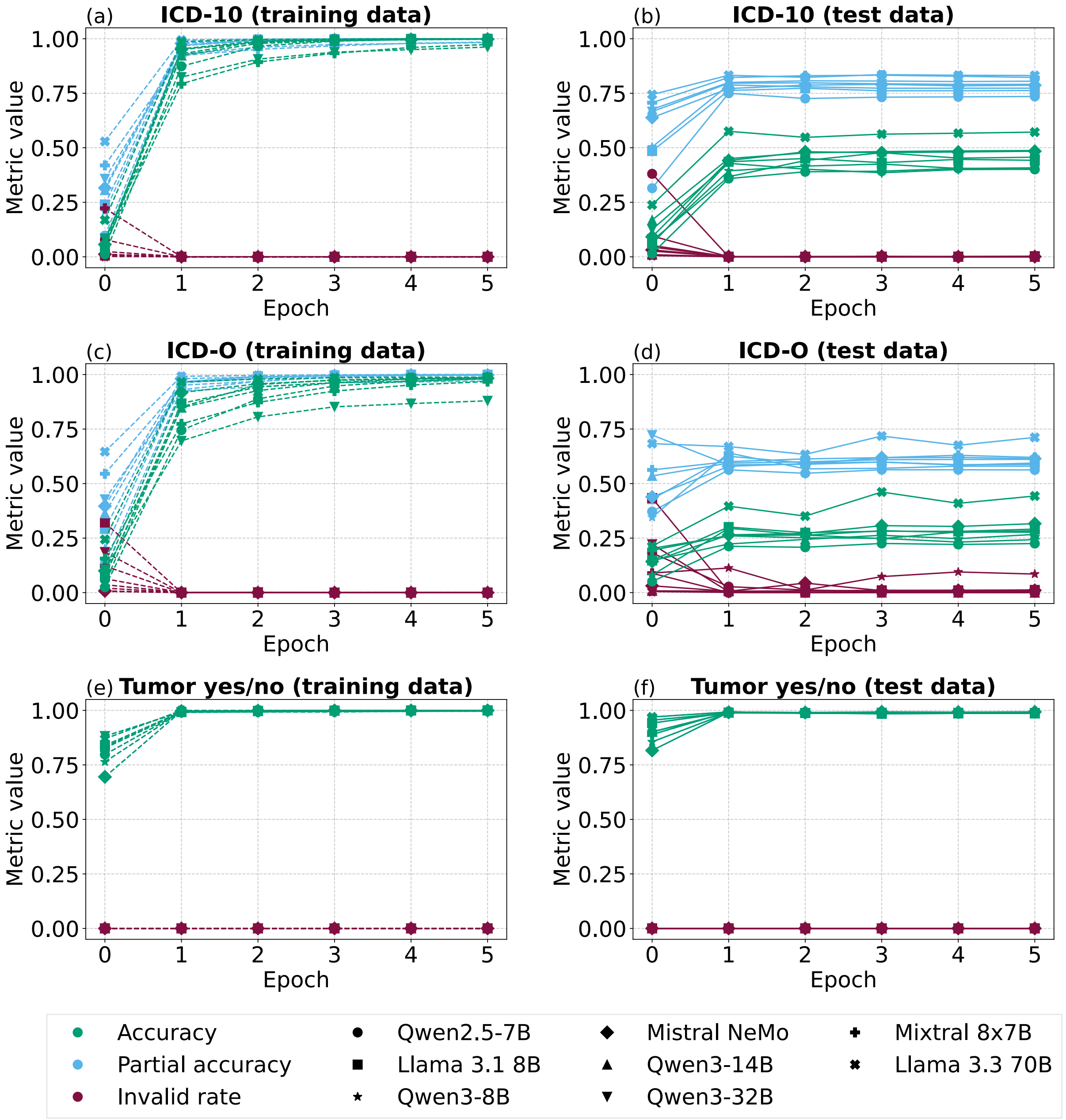

**Figure 2.** Performance of base models (epoch 0) and fine-tuned models (epochs 1-5) on the training dataset (panels a, c, e) and the test dataset (panels b, d, f). The base models and the fine-tuned models are evaluated after each training epoch with respect to the accuracy (rate of completely correct ICD-10 code, correct ICD-O topography code, or correct recognition as tumor diagnosis), partial accuracy (rate of correct first three characters of the ICD-10 or ICD-O topography code), and the rate of invalid, i.e., not interpretable, answers from the models. The accuracy and partial accuracy are measured relative to the proportion of valid answers. An HTML-based, interactive version of the plot is available as Supplementary Figure S1. There it is possible to further disentangle the curves. Additionally, the absolute accuracy values, counting invalid answers as false answers, can be compared there.

For ICD-10 coding, the fine-tuning improved the accuracy by around 40 percentage points for most models and at least by 31.9 percentage points for Qwen3-14B. The values of accuracy and partial accuracy were calculated among valid answers only to distinguish knowledge of correct codes from knowledge about formatting the answers in the desired way. This is especially relevant for the base models. After fine-tuning, the rate of invalid answers, e.g., codes that could not be interpreted as ICD codes via regular-expression based pattern matching of the answers, was reduced to near zero for almost all of the best fine-tuned versions of the models.

The additional exploratory evaluation on the external HANCOCK dataset showed an even more pronounced improvement in the ICD-10 coding task compared to the primary test dataset. Detailed results for this additional analysis are provided in Appendix B.

The performance improvements reported here are based on zero-shot prompts for both fine-tuned and base models. To assess how the magnitude of the fine-tuning effect compares with performance gains achievable through few-shot prompting, we additionally evaluated the base models using three-shot and five-shot prompting (Supplementary Table S5). Few-shot prompting improved coding performance for most base models, although the effect varied between models and metrics. For exact ICD-10 coding, employing five-shot prompting changed accuracy by −1.22 to +4.30 percentage points compared with zero-shot prompting. The differences for ICD-O topography coding were in a similar range. Compared to the improvements observed with few-shot prompting, fine-tuning produced substantially larger improvements for both ICD-10 and ICD-O across the evaluated models. These results indicate a more substantial and consistent benefit of instruction tuning for ICD code prediction than of few-shot prompting alone.

In general, ICD-O topography coding remained more challenging than ICD-10 coding across the evaluated models. Producing well-formed ICD-O codes was more difficult for the models, which can also be seen in a higher rate of malformed answers for the base models and for some fine-tuned training checkpoints. These invalid responses for ICD-O mostly consisted of morphology codes or completely invented codes that are an invalid mashup of topography and morphology codes or a similarly looking combination of characters and numbers.

To further characterize differences in coding performance, we performed an exploratory analysis across broader ICD-10 and ICD-O topography code groups. Considerable variation was observed between code groups. For ICD-O topography coding, some groups showed a pronounced difference between exact and partial accuracy. For example, the median exact accuracy across fine-tuned models was 0% for C49 (connective, subcutaneous and other soft tissues), whereas the corresponding partial accuracy was 74.02%, indicating that the broader topography category was frequently identified correctly while the exact subcode remained difficult to determine. A similar pattern was observed for C47 (peripheral nerves and autonomic nervous system). In contrast, performance was low for both exact and partial coding in groups such as C42 (Hematopoietic and reticuloendothelial systems) and C48 (retroperitoneum and peritoneum). The most frequent ICD-O topography group, C44 (skin), accounted for 36.51% of the test dataset and achieved a median exact accuracy of only 31.57%, suggesting that this frequent category also contributed substantially to the overall number of ICD-O coding errors. Detailed results for ICD-10 and ICD-O code groups are provided in Supplementary Table S7 and Table S8.

The manually re-annotated subset was additionally used to examine the relationship between code derivability and model performance (Supplementary Table S9). As expected, exact coding accuracy was higher on cases classified as fully derivable than on the complete test dataset for all fine-tuned models, for both ICD-10 and ICD-O topography coding. For ICD-O topography coding, partial accuracy remained relatively high for many models even among cases for which the exact localization was classified as only partially derivable from the diagnosis text. This indicates that the broader three-character topography category could often still be identified despite insufficient information for determining the exact subcode. At the same time, substantial coding errors also remained among cases classified as fully derivable, indicating that missing localization information alone does not account for the lower ICD-O performance. A similar pattern was observed for ICD-10, where the fine-tuned models correctly identified the three-character category in 92.36–100% of cases classified as only partially derivable.

**Table 4**. Comparison of the performances of the base models vs. the fine-tuned models on the test dataset. Only the values for the best training checkpoints (epochs) with respect to accuracy on the ICD-10 test dataset are shown. The accuracy (column "Acc.") / partial accuracy (column "P. acc") is calculated as the proportion of correct / partially correct answers regarding only valid answers, i.e. valid codes or valid yes/no answers, respectively. The proportion of invalid, i.e. uninterpretable, answers is reported separately (column "Inv."). The number X in "(best=X)" indicates the training epoch with the highest ICD-10 test accuracy for each model. The best accuracies are marked with bold font.

| | **ICD-10** | | | **ICD-O** | | | **Tumor diagnosis (y/n)** | |
|---|---|---|---|---|---|---|---|---|
| | **Acc.** | **P. acc.** | **Inv.** | **Acc.** | **P. acc.** | **Inv.** | **Acc.** | **Inv.** |
| **Qwen2.5-7B** | | | | | | | | |
| Base model | 1.43% | 31.46% | 38.05% | 5.09% | 37.17% | 18.17% | 90.16% | 0.00% |
| Trained (best=4) | 40.57% | 73.35% | 0.02% | 22.07% | 56.33% | 0.64% | 98.64% | 0.00% |
| **Llama 3.1 8B** | | | | | | | | |
| Base model | 5.86% | 48.48% | 5.21% | 14.86% | 42.85% | 43.35% | 94.19% | 0.01% |
| Trained (best=2) | 45.08% | 77.28% | 0.00% | 27.47% | 59.43% | 0.73% | 98.77% | 0.00% |
| **Qwen3-8B** | | | | | | | | |
| Base model | 6.69% | 50.35% | 2.64% | 8.02% | 34.52% | 8.98% | 85.57% | 0.00% |
| Trained (best=3) | 42.52% | 77.30% | 0.00% | 28.38% | 56.99% | 7.34% | 98.73% | 0.00% |
| **Mistral NeMo** | | | | | | | | |
| Base model | 9.17% | 63.84% | 3.17% | 14.26% | 43.87% | 3.12% | 81.62% | 0.00% |
| Trained (best=5) | 48.42% | 78.66% | 0.08% | 31.63% | 61.45% | 1.04% | **99.29%** | 0.00% |
| **Qwen3-14B** | | | | | | | | |
| Base model | 16.84% | 67.63% | 0.96% | 14.73% | 53.52% | 0.80% | 90.13% | 0.00% |
| Trained (best=5) | 48.72% | 80.54% | 0.00% | 26.69% | 61.09% | 0.00% | 98.71% | 0.00% |
| **Qwen3-32B** | | | | | | | | |
| Base model | 12.58% | 66.57% | 4.49% | 20.29% | 72.24% | 22.18% | 95.50% | 0.00% |
| Trained (best=1) | 42.87% | 79.59% | 0.01% | 26.05% | 58.61% | 0.42% | 98.83% | 0.00% |
| **Mixtral 8x7B** | | | | | | | | |
| Base model | 7.38% | 70.66% | 9.53% | 19.42% | 56.28% | 8.87% | 88.93% | 0.02% |
| Trained (best=3) | 47.75% | **83.27%** | 0.06% | 24.75% | 61.89% | 0.07% | 98.59% | 0.00% |
| **Llama 3.3 70B** | | | | | | | | |
| Base model | 23.81% | 74.41% | 0.57% | 21.35% | 68.33% | 0.51% | 96.95% | 0.00% |
| Trained (best=1) | **57.54%** | 83.26% | 0.04% | **39.64%** | **66.98%** | 0.11% | 99.26% | 0.00% |

Because accuracy was calculated among valid answers only, the ICD-O accuracy values of the base models must be interpreted together with the higher proportions of invalid ICD-O outputs. Thus, a higher accuracy among valid ICD-O answers does not necessarily indicate better overall ICD-O coding performance. Supplementary Figure S1 additionally shows the corresponding accuracy values when invalid outputs are counted as incorrect.

Tumor diagnosis recognition constitutes a binary classification task with a substantially smaller output space than ICD coding, which requires selection among hundreds of possible codes. Accordingly, the models achieved substantially higher accuracy on tumor diagnosis recognition than on ICD-10 and ICD-O coding. Interestingly, the base models were not able to get the tumor diagnosis recognition task right in all cases. On the training data, the errors of the base models consisted almost entirely of false-negative classifications, whereas false positives were rare (<0.4%). False-negative rates were lower on the test data and decreased to approximately 1% after fine-tuning, corresponding to an accuracy of 99% on training and test data. Most of the remaining errors in the test data set involved diagnosis descriptions containing (non-standard) abbreviations that were not recognized as tumor diagnoses, e.g., "PECA Unterkiefer links" (similar to "SCC left mandible", but with PECA being an unusual abbreviation for squamous cell carcinoma). A detailed breakdown of the false positive and false negative rates is provided in Supplementary Table S6.

Table 4 summarizes the performance of the best performing training checkpoints of the models. The performance values of the best performing fine-tuned versions are also shown in Figure 3. Values for weighted F1 scores are reported in Supplementary Table S4. The values shown in Table 4 (and Supplementary Table S4) reflect the single best fine-tuned checkpoint for each model, using ICD-10 accuracy as the selection criterion. The reported performance values therefore correspond to the fine-tuned model checkpoints that were selected for publication on HuggingFace. Supplementary Table S3 shows how the performance changes when the checkpoint with the highest ICD-O topography accuracy is selected instead. For three of the eight models, the same checkpoint was the best for both ICD-10 and ICD-O exact accuracy. For the remaining models, differences were generally small, with the most notable improvement being a 6.48 percentage-point increase in exact ICD-O accuracy for Llama 3.3 70B.

### 3.3. *Model size and performance*

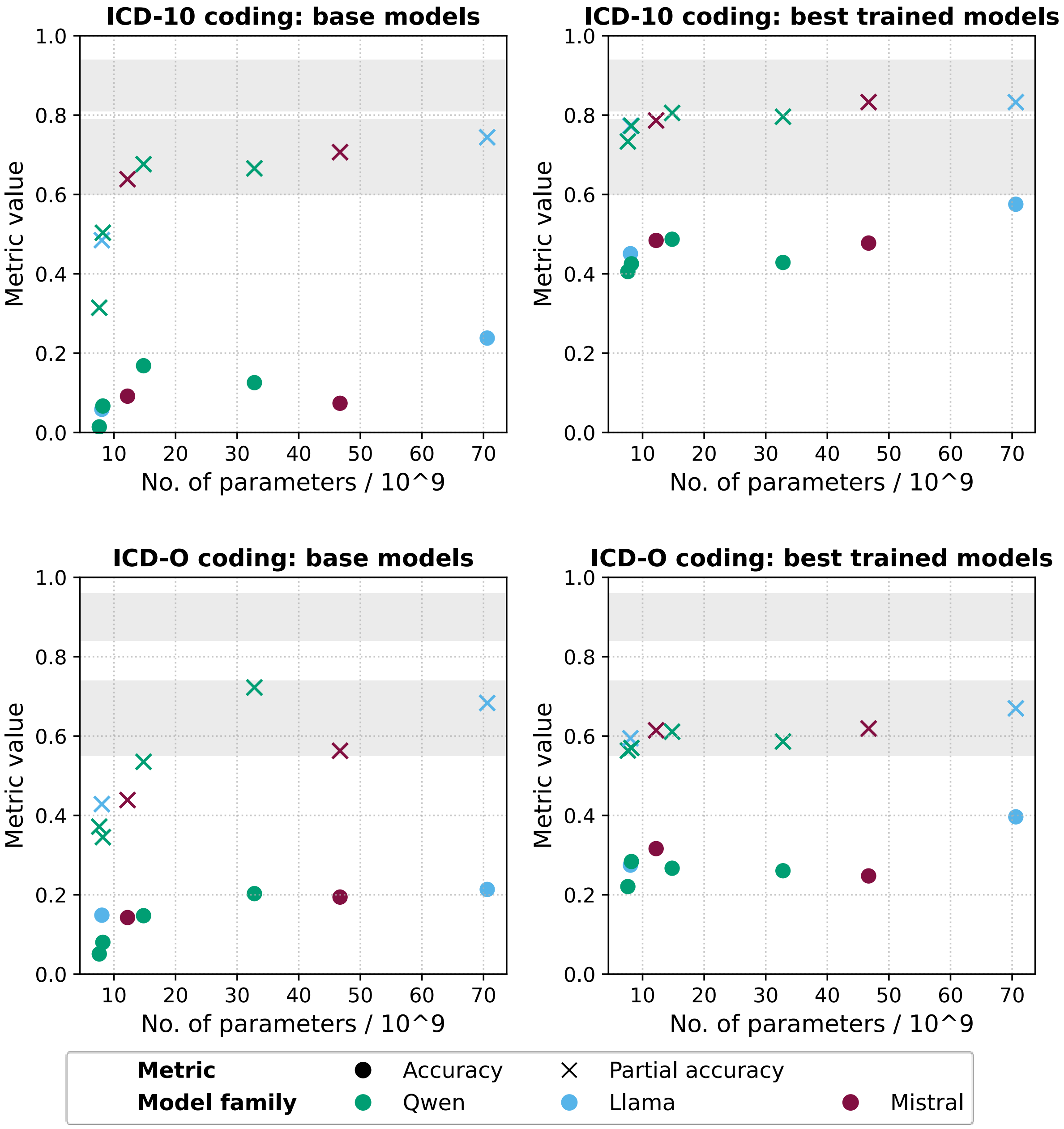

**Figure 3.** Performance of base models and best trained models for ICD-10 and ICD-O topography coding on the test dataset. The shaded grey bands represent 95% confidence intervals for the estimated upper limits of accuracy (lower grey band) and partial accuracy (upper grey band), which are derived from the data quality analysis.

Figure 3 is designed to facilitate a deeper look at how model size relates to baseline performance and the ability to improve via fine-tuning. The graphical display suggests a generally positive association between the number of parameters and accuracy, both for the base models and for the best fine-tuned models. The partial accuracy shows a sharp increase up to 14B parameters, particularly pronounced in ICD-10 coding. In contrast, the best fine-tuned models follow a more consistently upward trend. Fine-tuning reduces the initial performance gap between smaller and larger models, although larger fine-tuned models maintain a small advantage. For partially matching ICD-10 codes, the performance of the best trained models reached the range indicated by the confidence band for the maximally attainable accuracy. Given the uncertainty associated with the estimated upper bound, it remains unclear how much room is left for further improvement of the partial accuracy in the two strongest models, Llama 3.3 70B and Qwen3-14B. Compared to partial ICD-10 code matches, the performance gap is larger for exact ICD-10 code matches, and even greater for both exact and partial matches in ICD-O coding, indicating substantial potential for improvement in these tasks.

### 3.4. *Effect and computational cost of thinking mode*

A further point of investigation was the use of the thinking mode in the Qwen3 models. For this purpose, Figure 4 compares the performance of the Qwen3 base models with and without thinking and the best fine-tuned version of the models. (The combination of fine-tuning and thinking mode was also tested. The results are, however, not shown in Figure 4, as our experiments showed that the models do not employ the thinking process after fine-tuning for answering the questions.) Figure 4 shows that the accuracy did not improve when the thinking mode was activated. On the contrary, the accuracy often even decreased compared to the base model in non-thinking mode. Interestingly, the partial accuracy improved at least slightly when using the thinking process. In almost all cases, the fine-tuned models performed better than the base models with or without thinking. The only exception to this was Qwen3-32B, where the fine-tuned model had a better performance with respect to the accuracy but a worse performance in the partial accuracy.

After fine-tuning, the rate of malformed answers for ICD-O questions dropped to near zero (0.42%) from relatively high baseline values of 18% (base model with thinking) and 22% (base model without thinking) for Qwen3-32B. The reduction in malformed outputs did not result in a corresponding improvement in coding accuracy. The thinking process generally did not seem to help in producing more valid codes. In some cases, the rate of invalid answers was even higher using the thinking mode, especially in the case of ICD-O coding.

The response time is also a crucial factor to consider. Generating answers in thinking mode was more than 100 times slower than providing direct answers (7.9–37 s/question vs. 0.05–0.19 s/question on average across datasets and models). The fine-tuned models responded even faster than the base models (0.04–0.15 s/question).

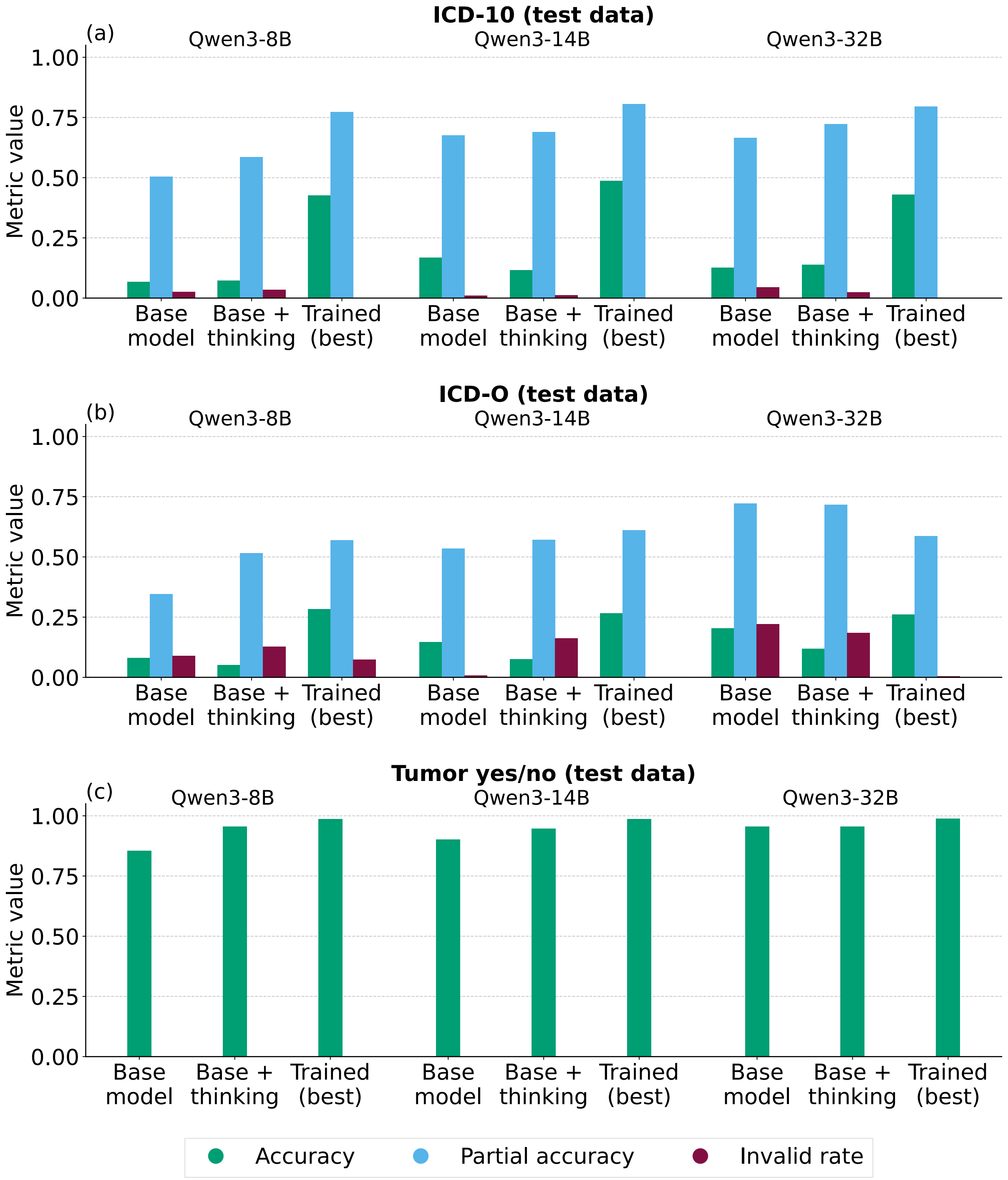

**Figure 4**. Comparing the ICD coding performance of Qwen3 models using thinking mode to the base models and the fine-tuned (trained) models. The values for the base models and the fine-tuned models were obtained using the complete test data set. Only 10% of the test data was evaluated using thinking mode, as answer generation in this mode is more than 100 times slower. The combination of trained models and thinking is not shown as the models do not use the thinking process after the fine-tuning.

## 4. Discussion

### *4.1. Limitations*

A central limitation of this study is the limited information contained in the tumor diagnosis descriptions in the real-world test dataset. The short free-text diagnosis descriptions were seen by the LLMs in isolation, whereas human coders have access to the clinical context detailed in doctors' notes and other types of medical documents. As a result, it is not guaranteed that codes can be derived from the short descriptions alone. We tried to quantify this constraint of incomplete information (see Section 2.3) and showed that only about 70% of ICD-10 codes and 65% of ICD-O codes are fully derivable, setting empirical upper bounds for the achievable accuracy below 100%. When employing LLMs for real-world coding tasks, providing richer contextual data could help models overcome the problem of insufficient information. This way, model performance may exceed what is reflected in our evaluation. The main result of this study, which is the demonstration that the coding accuracy can be improved substantially by using instructions from public data, is not undermined by the limitation of data quality.

There are more studies on automatic coding with LLMs that also used real-world datasets of short diagnosis descriptions [8,35]. While these studies did not report an estimation of an upper bound of achievable accuracy, we consider such an estimate very valuable, as it helps to interpret results more realistically and enables fairer comparisons of models and approaches across datasets.

The primary evaluation dataset originated from a single university cancer center and therefore reflects its local case mix and documentation practices. This limits the generalizability, as institution-specific jargon may affect model performance. Indeed, some of the remaining errors in tumor diagnosis recognition involved non-standard abbreviations that were not recognized by the models. At the same time, the models were fine-tuned exclusively on public standardized catalogue data rather than institution-specific clinical texts. The substantial improvements observed on the primary test dataset therefore indicate that such catalogue-based training can transfer to real-world clinical descriptions despite differences between formal catalogue terminology and local documentation language. Site-specific jargon and phrasing could be incorporated into additional instruction examples, providing a straightforward way to further adapt the approach to individual clinical settings.

The additional exploratory evaluation on the external HANCOCK dataset (see Appendix B) provides further evidence that the observed benefit of fine-tuning is not restricted to the primary test dataset. The improvement in ICD-10 coding performance was even more pronounced there compared to the main dataset, although its small size and its restriction to head and neck cancer diagnoses limit the conclusions that can be drawn from this analysis.

Using bigger models than the ones investigated here might improve the performance. Due to privacy concerns, hosting the models locally is necessary for a routine deployment [36]. For this reason, we only investigated models with a size that makes them usable with local resources inside the hospital in practice.

In this study, models of different sizes were compared with respect to their capability to be improved through fine-tuning. Figure 3 was created with the specific intention to correlate model performance with model size. There, a slight trend could be observed indicating that bigger models could keep a smaller advantage even after fine-tuning. However, this does not necessarily indicate that larger models must be superior to small models in all cases. More generally, model size alone is not expected to determine performance monotonically across different model families. Differences in model architecture, pretraining and post-training, quantization, and, for mixture-of-experts models, the distinction between total and activated parameters complicate direct comparisons based solely on nominal parameter count. Another reason that makes it difficult to interpret the effect of model size on the results is the implementation of the fine-tuning with low-rank adaptation (LoRA) [31] here. We applied the same LoRA configuration to all models, but this results in a different number of trainable parameters for different model sizes, as the affected matrices vary with model size. Consequently, larger models were given greater adaptation capacity during fine-tuning. One way to mitigate this issue would be to normalize the number of LoRA parameters across models. This could be done, e.g., by increasing the LoRA rank in smaller models so that their effective adaptation space becomes more comparable to that of larger models. However, such adjustments introduce their own trade-offs: higher ranks in smaller models increase training cost and may also raise the risk of overfitting. This could shed light on the influence of the number of adjusted parameters as a confounding factor in the model comparison but it would not necessarily reflect optimal fine-tuning practice.

Computational constraints prevented a systematic optimization of the training hyperparameters for the individual models. The smooth and rapidly decreasing loss curves observed during training (see Figure 1) indicate that the chosen hyperparameters were generally adequate for enabling the learning process. However, training instabilities occurred for Qwen3-14B and Qwen3-32B. This suggests that the chosen hyperparameters may not have been equally suitable for all models and may explain why Qwen3-32B performed below expectations for its size.

The computational constraints that prevented systematic optimization of the training hyperparameters also limited the scope for additional optimization experiments. In particular, we did not systematically investigate how the composition of the instruction-tuning dataset affects coding performance, for example by comparing different instruction formulations or different amounts of training data. Such experiments would address the additional question of how to optimize instruction design and data efficiency, whereas the objective of the present study was to evaluate a fixed training-data construction strategy using the available relevant catalogue information. Future studies could systematically vary these factors to determine which aspects of instruction-data construction contribute most strongly to coding performance.

The limitations discussed above highlight the importance of considering both data-related and model-related factors when interpreting results. Nevertheless, the demonstrated potential of instruction-based fine-tuning to enhance coding performance in LLMs remains unaffected.

### 4.2. *ICD-10 coding vs. ICD-O topography coding*

The models consistently performed worse on ICD-O topography coding compared to ICD-10 coding, although the relevant categories in ICD-O are fewer than in ICD-10. ICD-O topography coding may therefore be considered a less complex classification task if the localization information is given. Figure 3 shows that ICD-O performance falls short even when considering the lower theoretical ceiling imposed by data quality. This suggests that an important limiting factor is insufficient ICD-O knowledge acquired during model training, rather than data quality alone. The lower familiarity with ICD-O may result from limited exposure during pretraining, fine-tuning, or both. The relative contributions of pretraining exposure and fine-tuning sample size cannot be disentangled, as the extent to which ICD-O was represented in the pretraining data of the base models is unknown.

The analysis of ICD-O subgroups further showed that the limitations of ICD-O coding differed across anatomical sites (see Supplementary Table S8). For some code groups, partial accuracy was substantially higher than exact accuracy, suggesting that the models frequently identified the broader anatomical region but struggled to determine the specific subsite. In contrast, other groups showed low performance even at the three-character category level. Limited performance was also observed for the highly prevalent C44 category, which is therefore likely to contribute substantially to the overall number of ICD-O coding errors.

Together with the results from the manually re-annotated subset (see Supplementary Table S9), these findings indicate that insufficient localization information can limit exact topography coding but does not fully explain the remaining differences in performance between ICD-O code groups. Variation between code groups was also observed for ICD-10 coding (Supplementary Table S7), including substantial differences between exact and partial accuracy in some groups. However, the highly prevalent C43–C44 group (melanoma and other malignant neoplasms of the skin) showed relatively high performance. Consequently, the uneven code distribution influences the aggregate ICD-10 and ICD-O results in different ways.

Beyond these differences in particular subgroups, the generally lower ICD-O performance may also reflect generally weaker prior knowledge of the coding system in the base models. The likely reason for this is that the ICD-O-3 standard is less represented in general training corpora than ICD-10. This is also reflected by the fact that the base models showed systematic difficulty in distinguishing the concept of localization (ICD-O topography) from histological detail (ICD-O morphology). The frequent confusion between topography and morphology codes in the malformed outputs further supports this interpretation. The inclusion of morphology codes alongside topography codes in the training data was intended to help the models distinguish between the two code types. The training succeeded here, as the number of malformed answers was reduced to near zero for almost all models. Still, the models were less able to generalize ICD-O coding to real-world diagnosis descriptions after fine-tuning compared to ICD-10 coding. This was likely the case because the ICD-O catalogue used for training offers fewer synonyms and linguistic variations than the Alpha-ID catalogue for ICD-10. As a result, the models could not improve as much as they could for ICD-10 coding.

In summary, ICD-O performance lags behind not only because of the slightly lower performance ceiling set by data quality, but likely more because of weaker prior knowledge and lower training coverage. Future work should therefore focus on broadening the diversity of ICD-O coding examples. Targeted instruction-based training data for ICD-O coding should strengthen the understanding of concepts of localization and morphology, as well as the distinction and linkage between the concepts.

### 4.3. *Effect of employing the thinking mode on coding performance*

The Qwen3 thinking mode did not improve exact ICD coding accuracy and increased response times by more than two orders of magnitude. Although partial accuracy improved slightly in some settings, this pattern suggests that the reasoning process sometimes moved predictions toward the correct code category without reliably identifying the exact subcategory. Moreover, the reduction in malformed outputs did not consistently translate into improved coding accuracy, indicating that learning to produce formally valid codes is distinct from learning to select the correct code. For this task, targeted instruction-based fine-tuning was therefore more effective and substantially more computationally efficient than inference-time reasoning. The shorter response times of the fine-tuned models are also consistent with the training objective, which encouraged direct code-only answers rather than more verbose responses.

The fine-tuned Qwen3 models generally skipped the thinking process and produced direct answers. This behavior may be explained by the structure of the fine-tuning sequences. In the stream of tokens that the model learned to predict during training, the end-of-thinking token immediately followed the start-of-thinking token. This may have taught the models to omit intermediate reasoning steps when encountering these questions. Consequently, our results do not determine whether combining fine-tuning with training examples that explicitly preserve a reasoning process could improve performance. Nevertheless, the present findings indicate that unrestricted inference-time reasoning alone is not an efficient substitute for targeted domain-specific fine-tuning in this coding task.

### 4.4. *Comparison with additional methods for improving automatic coding*

An advantage of instruction-based fine-tuning is that it can address several related aspects of a coding task rather than training only a single classifier. The inclusion of complementary instructions may help models learn relevant conceptual distinctions, such as the distinction between tumor diagnoses, non-tumor diagnoses, and medical procedures. The clear improvement in tumor diagnosis recognition observed in our study suggests that the negative examples contributed to learning these distinctions. This improvement was notable because tumor diagnosis recognition was a comparatively simple task and the base models already achieved relatively high accuracy, yet still showed considerable room for improvement.

Beyond such differences in the scope of the learned task, automatic coding approaches also differ in how they derive the final code prediction. In our study, the textual output of the LLMs was used to infer the ICD codes. Other techniques for inferring codes make predictions based directly on the higher-level numeric representations ("embedding vectors") in neural networks.

In a study by Mittermeier et al. [37] concerning automatic ICD-10 coding of German diagnoses, the models GermanBERT and FlanT5 were trained on German radiology reports, and achieved accuracies of 58.8% and 65.2%, respectively, for predicting the ICD-10 category (i.e., "partial match" here) on the 100 most prevalent codes in the held-out test data set. For comparison, we achieved 83% partial match accuracy. Since we trained only on public data, our results are presumably more easily generalizable to other datasets of tumor diagnoses.

Kreuzthaler et al. [35] employed the BERT [38] based models medBERT.de and SapBERT-UMLS as embedding models for predicting ICD-10 codes (Austrian version) from short German diagnosis descriptions. Their large real-world dataset comprised over 20 million diagnosis/code pairs, which were split 80:10:10 into training, validation, and test sets. The coded diagnosis descriptions were transformed into embeddings with the two pretrained language models and used to classify diagnosis descriptions through similarity search. The achieved F1 scores for medBERT.de and SapBERT-UMLS on the test data were 0.86 and 0.87, respectively. If only public catalogue data were used to create the database of embedding vectors, Kreuzthaler et al. observed F1 scores of 0.55 and 0.61 on the test data. These values are in a similar range to the accuracy of 58% in our study. The work by Kreuzthaler et al. clearly demonstrates the potential of utilizing large-scale real-world data and embedding models.

In a very similar way, Böhringer et al. [39] conducted a study with three German hospitals, using a database of embedding vectors created with word2vec [40] to infer ICD-10 codes for ophthalmological diagnoses. Using data from the Alpha-ID catalogue and also real-world data from one hospital, they achieved accuracies for the exact matches of 69% and 69% on the data of two other hospitals, and partial accuracies of 86% and 91%. In comparison, the best accuracies achieved here were 58% for the exact match and 83% for the partial match. In contrast, our approach relied solely on the public catalogue and did not yet make use of tumor documentation data for training. It can be assumed that the performance can be improved further via instruction-based fine-tuning with more real-world data bringing in more variation. Furthermore, we used test data that was directly extracted from a database and conducted a data quality analysis but did not revise the data, while Böhringer et al. [39] tested the approach on diagnosis descriptions that were automatically extracted from 100 physicians' reports with regular expressions that were manually reviewed afterwards. Notably, the word2vec approach employed by Böhringer et al. is much more resource-efficient than the LLM approach, as it relies on a shallow single-layer neural network, which stands in contrast to the complex deep neural architectures used in LLMs. A hybrid strategy that combines such resource-parsimonious methods with LLMs for more complex cases may represent a promising direction for practical application.

Comparisons of performance are limited by differences in datasets and medical disciplines, and by the absence of a quantitative data quality analysis in the other studies. While accuracy is the most critical measure for clinical deployment, resource efficiency is also important for many real-world use cases. Notably, some of these methods are more resource-efficient than the way LLMs are employed here.

Table 5 provides an overview of the methodological differences between previous studies on ICD-coding of German diagnosis descriptions. Taken together, these studies demonstrate that different approaches to automated ICD coding have complementary strengths. Methods based on embedding vectors and large collections of real-world coded data can achieve high performance, but they require access to extensive clinical datasets and are often optimized for specific settings. In contrast, our approach provides a complementary pathway for improving automated coding by exploiting public data. Our unique contribution here is to demonstrate that publicly available classification catalogues can be transformed into instruction datasets to adapt open-weight LLMs for medical coding without requiring large institution-specific training datasets. Future approaches may combine the strengths of both paradigms by using efficient embedding-based retrieval methods together with instruction-tuned LLMs for handling more complex coding decisions.

**Table 5**. Comparison with other studies on ICD-coding of German diagnosis descriptions

| **Study** | **Clinical specialty** | **Training data** | **Primary metric** | **Method** |
|---|---|---|---|---|
| Böhringer et al. (2024) [39] | Ophthal-mology | ICD-10 coded short diagnosis descriptions from university hospital Freiburg, augmented with Alpha-ID | Accuracy | Semantic similarity search with word2vec |
| Mittermeier et al. (2024) [37] | Radiology | MRI reports with unique ICD-10 codes | Accuracy | Classification with FlanT5, GermanBERT |
| Kreuzthaler et al. (2025) [35] | None | ICD-10 coded short diagnoses from ten Austrian hospitals | Weighted F1 score | Semantic similarity search with medBERT.de and SapBERT |
| Lenz et al. (2025) [3] | Urologic oncology | None (only evaluation of ICD-10 coding) | Accuracy | Zero- and few-shot prompting of open-weight LLMs |
| This study | Oncology | Instructions created from public coding catalogues | Accuracy (based on well-formed answers) | Instruction-based fine-tuning of open-weight LLMs |

One method for combining embedding models and LLMs is retrieval-augmented generation (RAG) [41], where the embedding vectors are used to retrieve additional information from a database, such as a database of descriptions and ICD codes. Then the retrieved information, e.g., possible ICD codes and their descriptions, can be used to improve the queries for the LLMs and give them more information to reason about. Such a RAG approach has already been tested with the American coding system ICD-10-CM and English clinical notes from an emergency department [42]. Combining fine-tuned LLMs and RAG could further improve the performance compared to RAG that employs the basic pre-trained models.

## 5. Conclusions

We presented an approach for fine-tuning LLMs with question-answer pairs generated from German-language catalogues covering ICD-10 and ICD-O-3 localization coding. The main result of this study is that instructions crafted from public catalogue data can greatly enhance the coding abilities of current language models. This indicates that the potential of public resources has not been fully exploited for training open-weight LLMs and that there still is a considerable potential for improving the models with respect to specific medical documentation tasks by fine-tuning them with targeted instructions. Using public and standardized data for training allows for an easier generalization across different sites. The resulting models can also be directly re-used and serve as a foundation for site-specific training. Furthermore, the approach is also transferable to other languages and similar catalogues. Our complete dataset of the question-answer pairs and the best-performing training checkpoints of the fine-tuned models are available on Hugging Face[1].

We systematically analyzed the data quality as a decisive factor for judging LLM performance. Because routine diagnosis texts often lack essential details, perfect accuracy cannot be expected. By estimating realistic performance ceilings, we aimed to provide a fair interpretation of results. Our findings suggest that similar evaluations of coding short-form textual descriptions should be accompanied by an assessment of the completeness of the information in the texts to enable a more meaningful comparison with other studies.

We further analyzed the association between model size and model performance before and after training. Larger models achieve slightly better baseline results and retain an advantage after fine-tuning. Performance showed an overall positive association with model size, but much of the initial gap between smaller and larger base models can be closed through instruction-tuning. Adjusting fine-tuning strategies or diversifying the instruction set could further reduce this difference. Exploiting the reasoning mode in the Qwen3 models did not yield clear performance benefits in this use case, while it drastically increased the runtime. Nevertheless, the reasoning did steer predictions closer to the right categories, suggesting potential for integrating targeted instructions with guided reasoning in the future.

We believe that future progress will come from integrating LLMs with supporting methods, but also from improving the domain knowledge of LLMs, which was the focus of this study. Our findings demonstrate that there is an untapped potential for improving the coding performance of open-weight LLMs via instruction-based fine-tuning with publicly available data. The creation of more comprehensive, open instruction datasets, similar to the one in the approach here, is a promising way to advance the capabilities of LLMs with respect to medical documentation tasks.

## Statements and Declarations

### *Ethics approval and consent to participate*

The study was conducted in accordance with the principles of the Declaration of Helsinki. The requirement for a formal ethics review was waived by the responsible ethics commission (Ethikkommission der Landesärztekammer Rheinland-Pfalz) on 06 September 2024 because only short diagnosis descriptions and codes without identifying information were used for this study. Informed consent was waived by the institutional data protection office according to §

[1] https://huggingface.co/datasets/stefan-m-lenz/ICDOPS-QA-2024

37 (1) (3) of the State Hospital Act of Rhineland-Palatinate (Landeskrankenhausgesetz Rheinland-Pfalz) because only de-identified data were used in this study.

*Consent for publication*

Not applicable

*Data availability*

The training dataset and the best-performing checkpoints of the fine-tuned models are available from https://huggingface.co/datasets/stefan-m-lenz/ICDOPS-QA-2024.

*Funding*

This research has been supported by the Federal Ministry of Research, Technology and Space (BMFTR) in Germany in the project "Digitaler FortschrittsHub Gesundheit – DECIDE" (FKZ 01ZZ2106A) of the German national medical informatics initiative. The project aims to facilitate data transfer and interoperability among different stakeholders in the healthcare sector and to provide decision support based on structured patient information.

*Competing interests*

The authors declare that they have no competing interests relevant to the content of this article.

*Authors' contributions*

S.L. conceived the study, designed the experiments and wrote the manuscript. L.O.R. and S.L. carried out the experiments, and created the tables and plots. T.K. provided the data from the tumor documentation system, and G.V. performed the data extraction. A.U. and F.A. created the annotations for the data quality analysis, and A.U. analyzed them. F.A. also contributed to additional analyses and interpretation of the results during manuscript revision. T.P. provided the funding and supervised the work. All authors read and approved the final version of the manuscript.

# Appendix A

Appendix A contains the supplementary tables and figures referenced in the main manuscript, with the exception of the interactive Figure S1, which is contained in a separate HTML file.

**Table S1.** Distribution of ICD-10 codes in the test dataset. The first column shows the three-character categories that appear in the test data, ordered by frequency. The second column shows the number of different complete codes appearing in the data. The last two column show the absolute and relative frequency of the category.

| ICD-10 category | Distinct codes | Count | Percentage (%) |
|---|---|---|---|
| C44 | 8 | 393 | 19.42 |
| C43 | 7 | 288 | 14.23 |
| C34 | 6 | 108 | 5.34 |
| C83 | 5 | 87 | 4.30 |
| C61 | 1 | 78 | 3.85 |
| C64 | 1 | 57 | 2.82 |
| C67 | 7 | 53 | 2.62 |
| C71 | 9 | 50 | 2.47 |
| C15 | 5 | 45 | 2.22 |
| C80 | 1 | 42 | 2.08 |
| D03 | 6 | 42 | 2.08 |
| C49 | 7 | 39 | 1.93 |
| C16 | 5 | 37 | 1.83 |
| D32 | 2 | 37 | 1.83 |
| D09 | 3 | 36 | 1.78 |
| C73 | 1 | 35 | 1.73 |
| C62 | 3 | 33 | 1.63 |
| C81 | 5 | 32 | 1.58 |
| C22 | 6 | 28 | 1.38 |
| C03 | 3 | 25 | 1.24 |
| C25 | 5 | 23 | 1.14 |
| C82 | 5 | 23 | 1.14 |
| D33 | 4 | 22 | 1.09 |
| D47 | 4 | 18 | 0.89 |
| D48 | 3 | 18 | 0.89 |
| C32 | 3 | 17 | 0.84 |
| C84 | 4 | 17 | 0.84 |
| C17 | 5 | 16 | 0.79 |
| C18 | 9 | 16 | 0.79 |
| D43 | 5 | 16 | 0.79 |
| C02 | 2 | 14 | 0.69 |
| C24 | 3 | 14 | 0.69 |
| C09 | 2 | 13 | 0.64 |
| C13 | 5 | 13 | 0.64 |
| D46 | 4 | 13 | 0.64 |
| C04 | 3 | 12 | 0.59 |
| D35 | 1 | 11 | 0.54 |
| C10 | 4 | 9 | 0.44 |
| C30 | 1 | 9 | 0.44 |
| C69 | 4 | 9 | 0.44 |
| C01 | 1 | 8 | 0.40 |
| C05 | 3 | 8 | 0.40 |
| C20 | 1 | 8 | 0.40 |
| D07 | 2 | 8 | 0.40 |
| C41 | 2 | 7 | 0.35 |
| C48 | 1 | 7 | 0.35 |
| C66 | 1 | 7 | 0.35 |
| C21 | 2 | 6 | 0.30 |
| C65 | 1 | 6 | 0.30 |
| C00 | 1 | 5 | 0.25 |
| C06 | 1 | 5 | 0.25 |
| C07 | 1 | 5 | 0.25 |
| C40 | 2 | 5 | 0.25 |
| C46 | 2 | 5 | 0.25 |
| C60 | 3 | 5 | 0.25 |
| C72 | 2 | 5 | 0.25 |
| C85 | 2 | 5 | 0.25 |
| C86 | 1 | 5 | 0.25 |
| C11 | 2 | 4 | 0.20 |
| C37 | 1 | 4 | 0.20 |
| C50 | 2 | 4 | 0.20 |
| C96 | 2 | 4 | 0.20 |
| D41 | 3 | 4 | 0.20 |
| C38 | 2 | 3 | 0.15 |
| C47 | 3 | 3 | 0.15 |
| D01 | 3 | 3 | 0.15 |
| D39 | 2 | 3 | 0.15 |
| D40 | 1 | 3 | 0.15 |
| D45 | 1 | 3 | 0.15 |
| C08 | 1 | 2 | 0.10 |
| C23 | 1 | 2 | 0.10 |
| C31 | 2 | 2 | 0.10 |
| C45 | 1 | 2 | 0.10 |
| C52 | 1 | 2 | 0.10 |
| C63 | 2 | 2 | 0.10 |
| C70 | 1 | 2 | 0.10 |
| C74 | 1 | 2 | 0.10 |
| C75 | 1 | 2 | 0.10 |
| D02 | 1 | 2 | 0.10 |
| C54 | 1 | 1 | 0.05 |
| C56 | 1 | 1 | 0.05 |
| C68 | 1 | 1 | 0.05 |
| D00 | 1 | 1 | 0.05 |
| D21 | 1 | 1 | 0.05 |
| D37 | 1 | 1 | 0.05 |
| D38 | 1 | 1 | 0.05 |
| D44 | 1 | 1 | 0.05 |
| Total (87) | 238 | 2024 | 100 |

**Table S2**. Distribution of ICD-O-3 topography codes in the test dataset. The first column shows the three-character categories that appear in the test data, ordered by frequency. The second column shows the number of different complete codes appearing in the data. The last two column shows the absolute and relative frequency of the category.

| ICD-O category | Distinct codes | Count | Percentage (%) |
|---|---|---|---|
| C44 | 9 | 739 | 36.51 |
| C34 | 6 | 112 | 5.53 |
| C77 | 7 | 83 | 4.10 |
| C61 | 1 | 82 | 4.05 |
| C67 | 8 | 80 | 3.95 |
| C71 | 10 | 74 | 3.66 |
| C64 | 1 | 60 | 2.96 |
| C49 | 8 | 54 | 2.67 |
| C42 | 2 | 47 | 2.32 |
| C15 | 5 | 45 | 2.22 |
| C16 | 6 | 43 | 2.12 |
| C80 | 1 | 42 | 2.08 |
| C70 | 2 | 38 | 1.88 |
| C62 | 3 | 37 | 1.83 |
| C73 | 1 | 36 | 1.78 |
| C22 | 2 | 31 | 1.53 |
| C72 | 7 | 31 | 1.53 |
| C03 | 3 | 25 | 1.24 |
| C25 | 5 | 24 | 1.19 |
| C18 | 10 | 21 | 1.04 |
| C32 | 2 | 19 | 0.94 |
| C09 | 2 | 17 | 0.84 |
| C17 | 5 | 17 | 0.84 |
| C41 | 4 | 16 | 0.79 |
| C65 | 1 | 14 | 0.69 |
| C24 | 3 | 14 | 0.69 |
| C02 | 2 | 14 | 0.69 |
| C75 | 2 | 14 | 0.69 |
| C04 | 3 | 13 | 0.64 |
| C69 | 5 | 12 | 0.59 |
| C13 | 5 | 12 | 0.59 |
| C40 | 3 | 12 | 0.59 |
| C01 | 1 | 11 | 0.54 |
| C60 | 5 | 10 | 0.49 |
| C48 | 1 | 10 | 0.49 |
| C20 | 1 | 9 | 0.44 |
| C66 | 1 | 9 | 0.44 |
| C30 | 1 | 9 | 0.44 |
| C10 | 4 | 9 | 0.44 |
| C05 | 3 | 8 | 0.40 |
| C50 | 2 | 7 | 0.35 |
| C38 | 4 | 7 | 0.35 |
| C21 | 2 | 6 | 0.30 |
| C07 | 1 | 6 | 0.30 |
| C06 | 1 | 5 | 0.25 |
| C00 | 1 | 5 | 0.25 |
| C11 | 3 | 5 | 0.25 |
| C37 | 1 | 4 | 0.20 |
| C31 | 3 | 3 | 0.15 |
| C56 | 1 | 3 | 0.15 |
| C68 | 1 | 3 | 0.15 |
| C47 | 3 | 3 | 0.15 |
| C08 | 1 | 3 | 0.15 |
| C23 | 1 | 2 | 0.10 |
| C63 | 2 | 2 | 0.10 |
| C52 | 1 | 2 | 0.10 |
| C74 | 1 | 2 | 0.10 |
| C54 | 1 | 1 | 0.05 |
| C26 | 1 | 1 | 0.05 |
| C55 | 1 | 1 | 0.05 |
| Total (60) | 183 | 2024 | 100 |

**Table S3**. Performance change of models when considering the best training checkpoints for ICD-O topography coding instead of the best checkpoints for ICD-10 coding. The table shows the differences to the values in Table 4. The accuracy (column "Acc.") / partial accuracy (column "P. acc") is calculated using the test dataset as the proportion of correct / partially correct answers regarding only valid answers, i.e., valid codes or valid yes/no answers, respectively. The proportion of invalid, i.e., uninterpretable, answers is reported separately (column "Inv."). The first row of numbers for each model shows the values for taking the best training checkpoints for ICD-O topography coding, while the second row shows the differences to the best epochs for ICD-10 coding with the respective fine-tuned model in percentage points. "Epoch X vs. Y" indicates the percentage point difference of the optimal epoch for ICD-O topography coding (X) to the optimal epoch for ICD-10 coding (Y). The largest absolute differences for each of the performance metrics across models are marked with bold font.

| | **ICD-10** | | | **ICD-O** | | | **Tumor diagnosis (y/n)** | |
|---|---|---|---|---|---|---|---|---|
| | **Acc.** | **P. acc.** | **Inv.** | **Acc.** | **P. acc.** | **Inv.** | **Acc.** | **Inv.** |
| **Qwen2.5-7B** | 39.26% | 73.20% | 0.00% | 22.55% | 56.35% | 0.58% | 98.55% | 0.00% |
| Epoch 3 vs. 4 | -1.31% | -0.15% | -0.02% | +0.48% | +0.02% | -0.06% | -0.09% | 0% |
| **Llama 3.1 8B** | 43.57% | 76.27% | 0.02% | 30.08% | 62.54% | 0.67% | 98.95% | 0.00% |
| Epoch 1 vs. 2 | -1.51% | -1.01% | +0.02% | +2.61% | +3.11% | -0.06% | +0.18% | 0% |
| **Qwen3-8B** | 39.43% | 78.81% | 0.04% | 29.54% | 64.05% | 11.29% | 98.93% | 0.00% |
| Epoch 1 vs. 3 | **-3.09%** | **+1.51%** | +0.04% | +1.16% | **+7.06%** | **+3.95%** | +0.20% | 0% |
| **Mistral NeMo** | 48.42% | 78.66% | 0.08% | 31.63% | 61.45% | 1.04% | 99.29% | 0.00% |
| Epoch 5 vs. 5 | 0% | 0% | 0% | 0% | 0% | 0% | 0% | 0% |
| **Qwen3-14B** | 48.72% | 80.54% | 0.00% | 26.69% | 61.09% | 0.00% | 98.71% | 0.00% |
| Epoch 5 vs. 5 | 0% | 0% | 0% | 0% | 0% | 0% | 0% | 0% |
| **Qwen3-32B** | 42.87% | 79.59% | 0.01% | 26.05% | 58.61% | 0.42% | 98.83% | 0.00% |
| Epoch 1 vs. 1 | 0% | 0% | 0% | 0% | 0% | 0% | 0% | 0% |
| **Mixtral 8x7B** | 45.26% | 82.81% | 0.16% | 27.84% | 62.98% | 0.06% | 98.76% | 0.00% |
| Epoch 4 vs. 3 | -2.49% | -0.46% | **+0.10%** | +3.09% | +1.09% | -0.01% | +0.17% | 0% |
| **Llama 3.3 70B** | 56.25% | 83.54% | 0.05% | 46.12% | 71.81% | 0.02% | 98.69% | 0.00% |
| Epoch 3 vs. 1 | -1.29% | +0.28% | +0.01% | **+6.48%** | +4.83% | -0.09% | **-0.57%** | 0% |

**Table S4**. Weighted F1 scores for base models and fine-tuned models. The best scores are marked with bold font.

| | ICD-10 Weighted F1 score | | ICD-O Weighted F1 score | |
|---|---|---|---|---|
| | **Complete match** | **Partial match** | **Complete match** | **Partial match** |
| **Qwen2.5-7B** | | | | |
| Base model | 1.41% | 33.64% | 6.17% | 43.75% |
| Trained (best=4) | 44.85% | 75.83% | 27.29% | 62.82% |
| **Llama 3.1 8B** | | | | |
| Base model | 4.69% | 47.34% | 14.02% | 47.10% |
| Trained (best=2) | 49.89% | 79.00% | 34.50% | 65.75% |
| **Qwen3-8B** | | | | |
| Base model | 6.77% | 50.42% | 8.56% | 38.70% |
| Trained (best=3) | 47.82% | 78.76% | 34.44% | 64.26% |
| **Mistral NeMo** | | | | |
| Base model | 8.93% | 64.43% | 13.71% | 46.82% |
| Trained (best=5) | 54.42% | 80.94% | 39.45% | 68.34% |
| **Qwen3-14B** | | | | |
| Base model | 16.73% | 66.08% | 16.55% | 59.60% |
| Trained (best=5) | 52.83% | 82.36% | 33.49% | 67.97% |
| **Qwen3-32B** | | | | |
| Base model | 13.73% | 66.66% | 20.42% | 74.72% |
| Trained (best=1) | 47.69% | 80.49% | 31.84% | 65.05% |
| **Mixtral 8x7B** | | | | |
| Base model | 8.25% | 72.38% | 20.54% | 62.53% |
| Trained (best=3) | 52.19% | 84.57% | 31.19% | 68.75% |
| **Llama 3.3 70B** | | | | |
| Base model | 24.75% | 74.19% | 25.32% | 72.52% |
| Trained (best=1) | **62.40%** | **85.17%** | **47.38%** | **73.15%** |

**Table S5.** Comparison of few-shot prompting to zero-shot prompting for the baseline models. This table shows the difference to the values in Table 4 of the main manuscript if three-shot or five-shot prompting were employed instead of zero-shot prompting, which was used in the main analysis. The largest absolute differences for each of the performance metrics across models are marked with bold font.

| | ICD-10 | | | ICD-O | | |
|---|---|---|---|---|---|---|
| | **Acc.** | **P. acc.** | **Inv.** | **Acc.** | **P. acc.** | **Inv.** |
| **Qwen2.5-7B** | | | | | | |
| 3-shot vs. 0-shot | +1.36% | +0.94% | **-22.98%** | +2.17% | +2.07% | -16.89% |
| 5-shot vs 0-shot | +1.24% | +1.06% | **-26.93%** | +2.86% | +2.89% | -17.58% |
| **Llama-3.1-8B** | | | | | | |
| 3-shot vs. 0-shot | +1.21% | **+9.03%** | -2.29% | **-2.32%** | +1.65% | **-27.29%** |
| 5-shot vs 0-shot | +1.90% | **+9.61%** | -2.00% | **-3.77%** | +1.11% | **-28.87%** |
| **Qwen3-8B** | | | | | | |
| 3-shot vs. 0-shot | +0.05% | -2.90% | -2.39% | -1.16% | +0.62% | -8.39% |
| 5-shot vs 0-shot | +0.99% | -1.91% | -2.39% | -0.19% | +2.86% | -8.63% |
| **Mistral-Nemo** | | | | | | |
| 3-shot vs. 0-shot | +2.02% | -4.63% | +14.72% | +1.81% | **+3.30%** | +0.93% |
| 5-shot vs 0-shot | +1.53% | -3.08% | +13.23% | +1.64% | **+5.16%** | +0.54% |
| **Qwen3-14B** | | | | | | |
| 3-shot vs. 0-shot | +0.60% | -0.66% | -0.12% | 0% | -0.88% | -0.75% |
| 5-shot vs 0-shot | +0.81% | -0.40% | -0.32% | +0.10% | -0.58% | -0.75% |
| **Qwen3-32B** | | | | | | |
| 3-shot vs. 0-shot | **+3.98%** | +2.55% | -1.77% | -0.75% | -2.41% | -16.00% |
| 5-shot vs 0-shot | **+4.30%** | +2.63% | -1.38% | +1.58% | -1.31% | -17.54% |
| **Mixtral-8x7B** | | | | | | |
| 3-shot vs. 0-shot | +0.76% | -2.35% | -1.82% | -0.87% | -0.90% | -1.56% |
| 5-shot vs 0-shot | +0.64% | -3.55% | -1.92% | -0.69% | -0.34% | -2.50% |
| **Llama-3.3-70B** | | | | | | |
| 3-shot vs. 0-shot | -1.11% | -1.71% | +0.42% | +2.00% | +1.91% | +0.87% |
| 5-shot vs 0-shot | -1.22% | -1.30% | +0.57% | +2.53% | +1.60% | +0.58% |

**Table S6**. Rate of false positives (FP) and false negatives (FN) in the tumor diagnosis recognition task for the different models. The training data for the tumor diagnosis recognition task contained 50% tumor diagnoses and non-tumor diagnoses. Since the test data contained only tumor diagnoses, only the false negative rate can be reported there. The false positives and false negative are only reported for valid answers. The values for the accuracy and the rate of invalid answers for the tumor diagnosis recognition task are duplicated from Table 4 as comparison here.

| | Training data | | | | Test data | | |
|---|---|---|---|---|---|---|---|
| | Acc. | Inv. | FP | FN | Acc. | Inv. | FN |
| **Qwen2.5-7B** | | | | | | | |
| Base model | 79.75% | 0.00% | 0.23% | 20.03% | 90.16% | 0.00% | 9.84% |
| Trained (best=4) | 99.91% | 0.00% | 0.04% | 0.05% | 98.64% | 0.00% | 1.36% |
| **Llama 3.1 8B** | | | | | | | |
| Base model | 83.94% | 0.00% | 0.31% | 15.74% | 94.19% | 0.01% | 5.82% |
| Trained (best=2) | 99.75% | 0.00% | 0.24% | 0.01% | 98.77% | 0.00% | 1.23% |
| **Qwen3-8B** | | | | | | | |
| Base model | 76.38% | 0.00% | 0.10% | 23.51% | 85.57% | 0.00% | 14.43% |
| Trained (best=3) | 99.90% | 0.00% | 0.02% | 0.08% | 98.73% | 0.00% | 1.27% |
| **Mistral NeMo** | | | | | | | |
| Base model | 69.50% | 0.00% | 0.03% | 30.46% | 81.62% | 0.00% | 18.38% |
| Trained (best=5) | 99.97% | 0.00% | 0.00% | 0.02% | 99.29% | 0.00% | 0.71% |
| **Qwen3-14B** | | | | | | | |
| Base model | 83.35% | 0.00% | 0.22% | 16.43% | 90.13% | 0.00% | 9.87% |
| Trained (best=5) | 99.97% | 0.00% | 0.01% | 0.02% | 98.71% | 0.00% | 1.29% |
| **Qwen3-32B** | | | | | | | |
| Base model | 88.19% | 0.00% | 0.19% | 11.61% | 95.50% | 0.00% | 4.50% |
| Trained (best=1) | 99.21% | 0.00% | 0.38% | 0.41% | 98.83% | 0.00% | 1.17% |
| **Mixtral 8x7B** | | | | | | | |
| Base model | 82.72% | 0.01% | 0.17% | 17.12% | 88.93% | 0.02% | 11.09% |
| Trained (best=3) | 99.29% | 0.00% | 0.05% | 0.66% | 98.59% | 0.00% | 1.41% |
| **Llama 3.3 70B** | | | | | | | |
| Base model | 86.95% | 0.00% | 0.38% | 12.67% | 96.95% | 0.00% | 3.05% |
| Trained (best=1) | 99.95% | 0.00% | 0.04% | 0.01% | 99.26% | 0.00% | 0.74% |

**Table S7.** Performance of base and fine-tuned models across ICD-10 code groups. Values are reported as median [minimum–maximum] across the eight evaluated models. The percentage in parentheses below the code group indicates the proportion of the primary test dataset belonging to the respective group. Accuracy, partial accuracy, and the proportion of invalid outputs are reported separately for the base models and the best selected model snapshots after fine-tuning.

| ICD-10 code groups | Base models | | | Trained models | | |
|---|---|---|---|---|---|---|
| | **Accuracy (%)** | **P. accuracy (%)** | **Invalid (%)** | **Accuracy (%)** | **P. accuracy (%)** | **Invalid (%)** |
| **C00-C14** (6.08%) | 6.12 [0.00-24.53] | 25.88 [0.00-53.86] | 0.47 [0.00-28.79] | 27.00 [17.28-38.69] | 58.13 [51.26-69.11] | 0.00 [0.00-0.20] |
| **C15-C26** (9.63%) | 19.57 [4.26-27.52] | 65.35 [27.64-70.60] | 0.83 [0.00-8.42] | 44.75 [40.90-53.42] | 76.30 [72.35-78.71] | 0.00 [0.00-0.04] |
| **C30-C39** (7.07%) | 28.98 [6.70-42.46] | 78.72 [50.86-87.41] | 1.98 [0.06-39.16] | 42.74 [33.74-52.04] | 81.76 [77.39-85.14] | 0.00 [0.00-0.00] |
| **C40-C41** (0.59%) | 1.81 [0.00-12.23] | 32.01 [16.20-52.78] | 2.78 [0.00-12.50] | 0.00 [0.00-2.78] | 41.67 [41.67-50.00] | 0.00 [0.00-0.00] |
| **C43-C44** (33.65%) | 2.98 [0.06-21.86] | 82.56 [58.71-95.15] | 7.87 [0.00-61.44] | 59.25 [45.03-70.17] | 87.65 [76.85-92.41] | 0.00 [0.00-0.12] |
| **C45-C49** (2.77%) | 6.94 [1.49-12.22] | 60.46 [30.94-71.88] | 4.09 [0.00-12.80] | 2.83 [1.79-4.32] | 71.35 [68.30-76.04] | 0.00 [0.00-0.00] |
| **C50** (0.20%) | 26.81 [0.00-48.94] | 49.47 [41.46-50.00] | 1.04 [0.00-14.58] | 50.00 [50.00-50.00] | 50.00 [50.00-50.00] | 0.00 [0.00-0.00] |
| **C51-C58** (0.20%) | 0.00 [0.00-25.00] | 25.00 [0.00-45.83] | 0.00 [0.00-29.17] | 58.33 [25.00-75.00] | 58.33 [25.00-75.00] | 0.00 [0.00-0.00] |
| **C60-C63** (5.83%) | 15.75 [0.22-67.44] | 73.70 [31.50-89.26] | 0.04 [0.00-35.66] | 69.28 [62.01-74.65] | 86.05 [82.20-92.02] | 0.00 [0.00-0.00] |
| **C64-C68** (6.13%) | 6.71 [1.94-14.19] | 81.52 [38.60-88.03] | 0.47 [0.00-27.22] | 46.00 [23.79-52.62] | 88.78 [85.62-89.65] | 0.00 [0.00-0.00] |
| **C69-C72** (3.26%) | 4.66 [0.00-13.04] | 63.27 [1.77-90.38] | 0.69 [0.00-36.74] | 10.80 [2.90-19.32] | 90.28 [85.10-93.81] | 0.00 [0.00-0.00] |
| **C73-C75** (1.93%) | 0.54 [0.00-72.01] | 80.13 [39.58-85.01] | 1.07 [0.00-39.53] | 75.96 [49.15-85.26] | 83.55 [79.49-86.97] | 0.00 [0.00-0.00] |
| **C76-C80** (2.08%) | 0.31 [0.00-66.53] | 28.49 [1.45-85.10] | 3.37 [0.99-10.71] | 7.44 [1.79-59.72] | 44.44 [25.40-71.63] | 0.00 [0.00-1.59] |
| **C81-C96** (8.55%) | 12.72 [0.62-47.08] | 54.24 [17.31-73.80] | 0.29 [0.00-37.96] | 57.84 [40.75-74.42] | 87.52 [82.37-93.45] | 0.00 [0.00-0.05] |
| **D00-D09** (4.55%) | 0.34 [0.00-3.99] | 2.48 [0.00-16.64] | 3.67 [0.00-46.92] | 43.43 [33.15-68.75] | 57.84 [54.44-80.25] | 0.00 [0.00-0.09] |
| **D10-D36** (3.51%) | 0.12 [0.00-42.25] | 5.12 [0.00-84.91] | 0.12 [0.00-20.66] | 20.08 [12.09-31.46] | 64.15 [55.05-81.69] | 0.00 [0.00-0.94] |
| **D37-D48** (4.00%) | 3.89 [0.00-18.93] | 19.00 [0.21-22.09] | 2.62 [0.00-39.20] | 28.04 [23.46-35.80] | 53.34 [44.44-60.91] | 0.00 [0.00-0.62] |

**Table S8.** Performance of base and fine-tuned models across ICD-O topography code groups. Values are reported as median [minimum–maximum] across the eight evaluated models. The percentage in parentheses below the code group indicates the proportion of the primary test dataset belonging to the respective group. Accuracy, partial accuracy, and the proportion of invalid outputs are reported separately for the base models and the best selected model snapshots after fine-tuning.

| ICD-O code groups | Base models | | | Trained models | | |
|---|---|---|---|---|---|---|
| | Accuracy (%) | P. accuracy (%) | Invalid (%) | Accuracy (%) | P. accuracy (%) | Invalid (%) |
| **C00-C14** (6.57%) | 8.23 [0.08-29.15] | 30.82 [0.47-52.97] | 1.84 [0.15-27.14] | 26.14 [18.65-36.77] | 61.38 [54.51-78.65] | 0.00 [0.00-5.26] |
| **C15-C26** (10.52%) | 21.65 [4.49-30.00] | 66.81 [34.80-73.44] | 1.78 [0.14-23.10] | 38.63 [32.20-49.91] | 74.42 [70.38-76.50] | 0.09 [0.00-4.41] |
| **C30-C39** (7.61%) | 19.59 [7.52-27.98] | 74.74 [63.49-86.57] | 2.40 [0.00-25.13] | 38.84 [32.34-53.05] | 81.15 [77.13-87.14] | 0.06 [0.00-1.88] |
| **C40-C41** (1.38%) | 1.75 [0.00-6.98] | 22.05 [0.00-38.21] | 14.11 [0.00-79.64] | 0.00 [0.00-1.44] | 31.89 [27.47-34.78] | 1.25 [0.00-42.86] |
| **C42** (2.32%) | 2.39 [0.00-24.65] | 15.34 [0.00-36.74] | 10.85 [0.64-77.87] | 0.00 [0.00-29.64] | 0.55 [0.00-39.87] | 1.91 [0.00-67.23] |
| **C44** (36.51%) | 3.23 [0.43-20.70] | 41.28 [19.88-85.69] | 10.28 [0.03-51.87] | 31.57 [22.22-56.91] | 50.53 [41.77-71.91] | 0.02 [0.00-3.63] |
| **C47** (0.15%) | 0.00 [0.00-0.00] | 0.00 [0.00-70.37] | 5.00 [0.00-70.00] | 0.00 [0.00-0.00] | 79.52 [66.67-100] | 0.00 [0.00-53.33] |
| **C48** (0.49%) | 0.00 [0.00-34.62] | 0.00 [0.00-37.18] | 24.00 [0.00-84.00] | 0.00 [0.00-0.00] | 0.00 [0.00-0.00] | 0.50 [0.00-30.00] |
| **C49** (2.67%) | 0.74 [0.00-3.77] | 37.84 [0.00-69.39] | 23.70 [2.59-90.00] | 0.00 [0.00-1.50] | 74.02 [60.11-81.48] | 0.83 [0.00-16.67] |
| **C50** (0.35%) | 17.50 [0.00-37.04] | 30.08 [26.79-45.45] | 15.00 [0.00-45.71] | 11.43 [0.00-29.27] | 28.57 [28.57-48.78] | 0.00 [0.00-41.43] |
| **C51-C58** (0.35%) | 8.63 [0.00-22.86] | 13.46 [0.00-31.43] | 7.86 [0.00-61.43] | 24.29 [5.71-55.71] | 55.71 [28.99-71.43] | 0.00 [0.00-1.43] |
| **C60-C63** (6.47%) | 49.51 [19.92-69.77] | 77.53 [42.48-89.03] | 1.30 [0.00-28.78] | 59.34 [43.13-80.76] | 88.89 [85.59-91.37] | 0.08 [0.00-1.37] |
| **C64-C68** (8.20%) | 32.73 [5.89-44.06] | 80.84 [39.51-88.21] | 1.45 [0.18-25.90] | 17.35 [7.24-42.53] | 86.61 [85.55-89.09] | 0.03 [0.00-15.18] |
| **C69-C72** (7.66%) | 11.25 [0.71-21.74] | 58.95 [4.18-74.04] | 14.13 [0.00-54.13] | 10.90 [4.86-13.87] | 75.65 [72.36-79.23] | 0.39 [0.00-34.32] |
| **C73-C75** (2.57%) | 22.63 [0.00-59.17] | 76.61 [0.79-86.93] | 4.81 [0.00-77.88] | 36.69 [22.88-79.42] | 81.10 [79.81-86.36] | 0.38 [0.00-6.92] |
| **C76-C80** (6.18%) | 5.05 [0.72-15.37] | 12.10 [4.18-25.91] | 8.12 [0.40-50.00] | 11.08 [8.13-16.44] | 16.07 [9.81-24.48] | 0.52 [0.00-30.88] |

**Table S9.** Performance of base models and fine-tuned models on the manually re-annotated subset of the test dataset. The tasks ICD-10 coding and ICD-O topography coding are considered here. The values for the accuracy ("Acc."), partial accuracy ("P. acc") and invalid rate ("Inv.") are shown separately for the subsets of codes that were deemed fully derivable or only partially derivable with respect to their ICD-10 or ICD-O topography codes. Analogously to Table 4, accuracy and partial accuracy are calculated only on the set of valid answers. The best accuracies are marked with bold font.

| | **ICD-10** | | | | | **ICD-O** | | | | |
|---|---|---|---|---|---|---|---|---|---|---|
| | **Fully derivable** | | | **Partially derivable** | | **Fully derivable** | | | **Partially derivable** | |
| | **Acc.** | **P. acc.** | **Inv.** | **P. acc.** | **Inv.** | **Acc.** | **P. acc.** | **Inv.** | **P. acc.** | **Inv.** |
| **Qwen2.5-7B** | | | | | | | | | | |
| Base model | 3.98% | 41.81% | 36.56% | 58.44% | 46.53% | 9.70% | 57.62% | 12.96% | 64.14% | 20.80% |
| Trained (best=4) | 55.93% | 77.88% | 0.00% | **100.0%** | 0.00% | 26.31% | 71.90% | 0.89% | 71.02% | 2.00% |
| **Llama 3.1 8B** | | | | | | | | | | |
| Base model | 11.80% | 59.62% | 3.05% | 63.64% | 8.33% | 28.65% | 54.68% | 46.22% | 73.89% | 37.20% |
| Trained (best=2) | 61.62% | 86.97% | 0.00% | 93.06% | 0.00% | 34.18% | 74.78% | 0.74% | 93.98% | 0.40% |
| **Qwen3-8B** | | | | | | | | | | |
| Base model | 8.58% | 59.76% | 0.82% | 81.95% | 7.64% | 11.11% | 33.41% | 6.67% | 29.82% | 8.80% |
| Trained (best=3) | 50.88% | 83.04% | 0.00% | **100.0%** | 0.00% | 34.52% | 72.68% | 6.44% | 69.70% | 7.60% |
| **Mistral NeMo** | | | | | | | | | | |
| Base model | 15.45% | 78.04% | 1.64% | 84.44% | 6.25% | 36.07% | 62.73% | 1.63% | 69.48% | 0.40% |
| Trained (best=5) | 63.03% | 83.74% | 0.00% | **100.0%** | 0.00% | 38.67% | 73.55% | 0.59% | 70.56% | 0.80% |
| **Qwen3-14B** | | | | | | | | | | |
| Base model | 29.11% | 78.52% | 0.00% | 90.91% | 8.33% | 23.28% | 70.90% | 0.74% | 69.35% | 0.80% |
| Trained (best=5) | 58.92% | 83.39% | 0.00% | **100.0%** | 0.00% | 36.07% | 77.26% | 0.00% | 70.80% | 0.00% |
| **Qwen3-32B** | | | | | | | | | | |
| Base model | 16.62% | 74.93% | 2.17% | 75.54% | 3.47% | 34.5% | 81.62% | 10.15% | 95.34% | 22.80% |
| Trained (best=1) | 65.49% | 87.79% | 0.00% | 92.36% | 0.00% | 33.41% | 74.96% | 0.00% | 68.00% | 0.00% |
| **Mixtral 8x7B** | | | | | | | | | | |
| Base model | 11.63% | 74.62% | 2.64% | 90.77% | 9.72% | 42.31% | 72.68% | 5.11% | 68.20% | 4.40% |
| Trained (best=3) | 62.15% | **88.56%** | 0.00% | **100.0%** | 0.00% | 31.63% | 75.41% | 0.00% | 72.80% | 0.00% |
| **Llama 3.3 70B** | | | | | | | | | | |
| Base model | 52.93% | 83.69% | 0.00% | 83.33% | 0.00% | 31.72% | **83.17%** | 0.52% | **98.37%** | 2.00% |
| Trained (best=1) | **68.54%** | 86.15% | 0.00% | **100.0%** | 0.00% | **52.41%** | 76.20% | 0.07% | 80.00% | 0.00% |

# Appendix B

This appendix presents an exploratory evaluation on a small external dataset to provide an additional assessment of generalizability beyond the primary test dataset.

*Rationale and dataset*

To provide an additional assessment of the generalizability of the catalogue-based fine-tuning beyond the local tumor documentation data used in the main analysis, we performed an exploratory evaluation on an independent publicly available dataset.

For this purpose, we used data from the HANCOCK (Head And Neck Cancer dataset) dataset [20]. HANCOCK is a monocentric real-world dataset comprising multimodal data from 763 patients with head and neck cancer treated at University Hospital Erlangen. The retrospective cohort covers patients diagnosed between 2005 and 2019 and includes structured clinical and pathological information, histopathological images, laboratory data, and German-language surgery reports. ICD-10-GM and corresponding short descriptions of the tumor diagnoses were extracted from the clinical documents as part of the original dataset preparation.

*Preparation of the external evaluation dataset*

For the present analysis, we used the “Text data” component of the HANCOCK dataset. Within this component, the subdirectory “icd_codes” contains individual text files with one or more short diagnosis descriptions and their corresponding ICD-10-GM codes. Each text-code assignment was treated as a separate entry in the external evaluation dataset.

Because the models had been fine-tuned using descriptions from the Alpha-ID catalogue, direct textual overlap between the external evaluation data and the catalogue-based training data could artificially inflate performance. We therefore compared the diagnosis descriptions from HANCOCK with the Alpha-ID descriptions used for training and excluded entries for which an identical textual description occurred in the training data. Additionally, we removed all overlap with the primary evaluation dataset by removing all entries that were present in the primary dataset.

After this filtering step, the external evaluation dataset contained 87 diagnosis descriptions representing 27 distinct ICD-10-GM subcodes across 13 three-character ICD-10-GM categories. The distribution of the remaining codes is shown in Table B1. It differs markedly from that of the primary test dataset, whose corresponding ICD-10 code distribution is shown in Supplementary Table S1, reflecting the focus of HANCOCK on head and neck cancers.

**Table B1**. Distribution of ICD-10 codes in the evaluation dataset derived from the HANCOCK data

| ICD-10 Code | Count | Distinct Subcodes | Percentage (%) |
|---|---|---|---|
| C10 | 16 | 4 | 18.39 |
| C13 | 14 | 4 | 16.09 |
| C32 | 12 | 3 | 13.79 |
| C02 | 11 | 2 | 12.64 |
| C09 | 11 | 3 | 12.64 |
| D37 | 7 | 1 | 8.05 |
| C05 | 5 | 3 | 5.75 |
| C04 | 4 | 2 | 4.60 |
| C06 | 3 | 1 | 3.45 |
| C11 | 1 | 1 | 1.15 |
| C14 | 1 | 1 | 1.15 |
| C18 | 1 | 1 | 1.15 |
| C80 | 1 | 1 | 1.15 |
| Total (13) | 87 | 27 | 100 |

*Model evaluation*

The external dataset was used exclusively for evaluation and was not included in model training or checkpoint selection. We evaluated the same base models and fine-tuned model checkpoints as in the primary analysis. ICD-10-GM coding questions were constructed and evaluated using the same procedure as for the primary test dataset. Exact accuracy, partial accuracy based on the three-character ICD-10-GM category, the proportion of invalid outputs, and weighted F1 scores were calculated according to the same procedures and definitions used in the main analysis. Because of the comparatively small size and restricted clinical domain of the external dataset, this analysis was considered an exploratory external evaluation rather than a separate validation study.

*External evaluation results*

Fine-tuning substantially improved ICD-10-GM coding performance for all eight models on the external dataset (see Table B2). Exact accuracy increased from 1.27–34.48% for the base models to 63.22–86.21% after fine-tuning, while partial accuracy increased from 8.75–72.41% to 79.31–95.40%. The largest absolute increase in exact accuracy was observed for Llama 3.1 8B, with an improvement from 4.60% to 77.01%. Llama 3.3 70B achieved the highest absolute performance after fine-tuning, with an exact accuracy of 86.21% and a partial accuracy of 95.40%. Weighted F1 scores showed the same overall pattern, with substantial improvements after fine-tuning. Invalid outputs were absent for all fine-tuned models.

The improvement after fine-tuning was more pronounced than in the primary test dataset, and the fine-tuned models also achieved higher absolute ICD-10 coding accuracies on the external dataset. However, these absolute performance values should not be interpreted as indicating superior general coding performance on external data. The HANCOCK-derived evaluation set is small and restricted primarily to head and neck cancer diagnoses, with a markedly different code distribution from the primary test dataset. The analysis therefore provides additional evidence that the benefit of catalogue-based fine-tuning extends to diagnosis descriptions from another institution, while the magnitude of the observed performance should be interpreted in the context of the restricted external dataset.

**Table B2.** ICD-10 coding performance of base models and trained models on the external evaluation dataset derived from HANCOCK. The best scores are marked with bold font.

| | **Accuracy** | **Partial accuracy** | **Invalid** | **Weighted F1 complete match** | **Weighted F1 partial match** |
|---|---|---|---|---|---|
| **Qwen2.5-7B** | | | | | |
| Base model | 1.27% | 8.86% | 9.20% | 1.27% | 6.44% |
| Trained (best=4) | 70.11% | 85.06% | 0.00% | 76.80% | 86.67% |
| **Llama 3.1 8B** | | | | | |
| Base model | 4.60% | 11.49% | 0.00% | 5.09% | 13.76% |
| Trained (best=2) | 77.01% | 93.10% | 0.00% | 81.70% | 93.68% |
| **Qwen3-8B** | | | | | |
| Base model | 3.45% | 27.59% | 0.00% | 3.45% | 20.30% |
| Trained (best=3) | 68.97% | 94.25% | 0.00% | 73.76% | 95.32% |
| **Mistral NeMo** | | | | | |
| Base model | 10.34% | 45.98% | 0.00% | 7.99% | 47.85% |
| Trained (best=5) | 63.22% | 80.46% | 0.00% | 67.66% | 83.11% |
| **Qwen3-14B** | | | | | |
| Base model | 14.94% | 49.43% | 0.00% | 13.85% | 40.89% |
| Trained (best=5) | 80.46% | 94.25% | 0.00% | 82.23% | 95.33% |
| **Qwen3-32B** | | | | | |
| Base model | 8.05% | 50.57% | 0.00% | 8.46% | 49.77% |
| Trained (best=1) | 78.16% | 94.25% | 0.00% | 82.82% | 95.31% |
| **Mixtral 8x7B** | | | | | |
| Base model | 17.86% | 50.00% | 3.45% | 19.82% | 46.41% |
| Trained (best=3) | 64.37% | 79.31% | 0.00% | 67.84% | 83.72% |
| **Llama 3.3 70B** | | | | | |
| Base model | 34.48% | 72.41% | 0.00% | 39.43% | 71.02% |
| Trained (best=1) | **86.21%** | **95.40%** | 0.00% | **88.06%** | **95.57%** |